\PassOptionsToPackage{dvipsnames}{xcolor}

\documentclass[sigconf,natbib=true,anonymous=false]{acmart} % 1. 取消匿名模式

\renewcommand\footnotetextcopyrightpermission[1]{} % 移除脚注版权
\setcopyright{none} % 设置版权为 none

\usepackage[most]{tcolorbox} % 加载 tcolorbox 及其所有库
\usepackage{algorithm}
\usepackage{algpseudocode}
\usepackage{multirow}
\usepackage{makecell}
\usepackage{listings}

\AtBeginDocument{%
  }
\providecommand{\appendixfloatnumbering}{}
\renewcommand\footnotetextcopyrightpermission[1]{}
\setcopyright{acmlicensed}
\copyrightyear{2018}
\acmYear{2018}
\acmDOI{XXXXXXX.XXXXXXX}
\acmConference[Conference acronym 'XX]{Make sure to enter the correct
  conference title from your rights confirmation email}{June 03--05,
  2018}{Woodstock, NY}

\acmISBN{978-1-4503-XXXX-X/2018/06}
\acmSubmissionID{584}

\begin{document}

\title{MMGraphRAG: Bridging Vision and Language with Interpretable Multimodal Knowledge Graphs}

%% --- 3. 更新作者信息 ---
\author{Xueyao Wan}
\affiliation{%
  \institution{Harbin Institute of Technology}
  \country{China}
}
\email{1185909349@qq.com}

\author{Hang Yu}
\affiliation{%
  \institution{Nanyang Technological University}
  \country{Singapore}
}
\email{hang023@e.ntu.edu.sg}
\begin{abstract}

Large Language Models (LLMs) suffer from hallucinations due to their static parametric knowledge. Retrieval-Augmented Generation (RAG) and GraphRAG mitigate this issue by incorporating external knowledge and structured reasoning over knowledge graphs (KGs). However, existing approaches remain largely text-centric, as constructing fine-grained multimodal knowledge graphs (MMKGs) with explicit cross-modal semantics remains challenging.
In this paper, we propose MMGraphRAG, a framework for building interpretable MMKGs that unify textual and visual knowledge. Our approach represents visual content as structured scene graphs and integrates them with textual KGs through a novel cross-modal entity linking method, SpecLink, which leverages spectral clustering to jointly model semantic similarity and graph structure. This design preserves explicit entities, relations, and reasoning paths across modalities, enabling structure-aware retrieval and generation.
To support evaluation, we introduce the CMEL dataset, a benchmark for fine-grained cross-modal entity alignment. Experimental results on CMEL demonstrate improved entity linking accuracy, while evaluations on DocBench and MMLongBench show that MMGraphRAG achieves superior performance and stronger robustness, particularly in complex multimodal reasoning scenarios.

\keywords{Multimodal Knowledge Graph \and Retrieval-Augmented Generation \and Cross-Modal Entity Linking}

\end{abstract}

\begin{CCSXML}
<ccs2012>
   <concept>
       <concept_id>10002951.10003317.10003347.10003348</concept_id>
       <concept_desc>Information systems~Question answering</concept_desc>
       <concept_significance>500</concept_significance>
       </concept>
   <concept>
       <concept_id>10002951.10003317.10003318.10003321</concept_id>
       <concept_desc>Information systems~Content analysis and feature selection</concept_desc>
       <concept_significance>500</concept_significance>
       </concept>
 </ccs2012>
\end{CCSXML}

\ccsdesc[500]{Information systems~Question answering}
\ccsdesc[500]{Information systems~Content analysis and feature selection}

\keywords{GraphRAG, Cross-Modal Entity Linking, Scene Graph}

\received{20 February 2007}
\received[revised]{12 March 2009}
\received[accepted]{5 June 2009}

\maketitle
\section{Introduction}

Large Language Models (LLMs) have achieved remarkable progress in natural language generation, yet hallucination, i.e., factual inconsistency, remains a fundamental limitation~\cite{DBLP:journals/corr/abs-2004-13637,DBLP:journals/corr/abs-2402-00253,10.1145/3703155}. This issue largely stems from their static parametric knowledge, which cannot be readily updated or grounded in domain-specific information~\cite{chu2025reducing,chen2025mrdragenhancingmedicaldiagnosis,xia2025mmedrag,10921633}. Retrieval-Augmented Generation (RAG) alleviates this limitation by incorporating external knowledge sources, while recent advances such as GraphRAG further enhance interpretability and reasoning by structuring retrieved knowledge as knowledge graphs (KGs)~\cite{edge2025localglobalgraphrag}.

However, existing GraphRAG approaches remain largely text-centric. In real-world scenarios, information is inherently multimodal, spanning text, images, and tables. Restricting retrieval and reasoning to textual KGs leads to incomplete knowledge grounding and limits the ability to capture cross-modal evidence~\cite{NEURIPS2023_47393e85}. A key challenge lies in constructing retrieval-oriented multimodal entity-relation graphs that preserve explicit entities, relations, and cross-modal links from heterogeneous documents.

\begin{figure}[h]
\centering
\includegraphics[width=\linewidth]{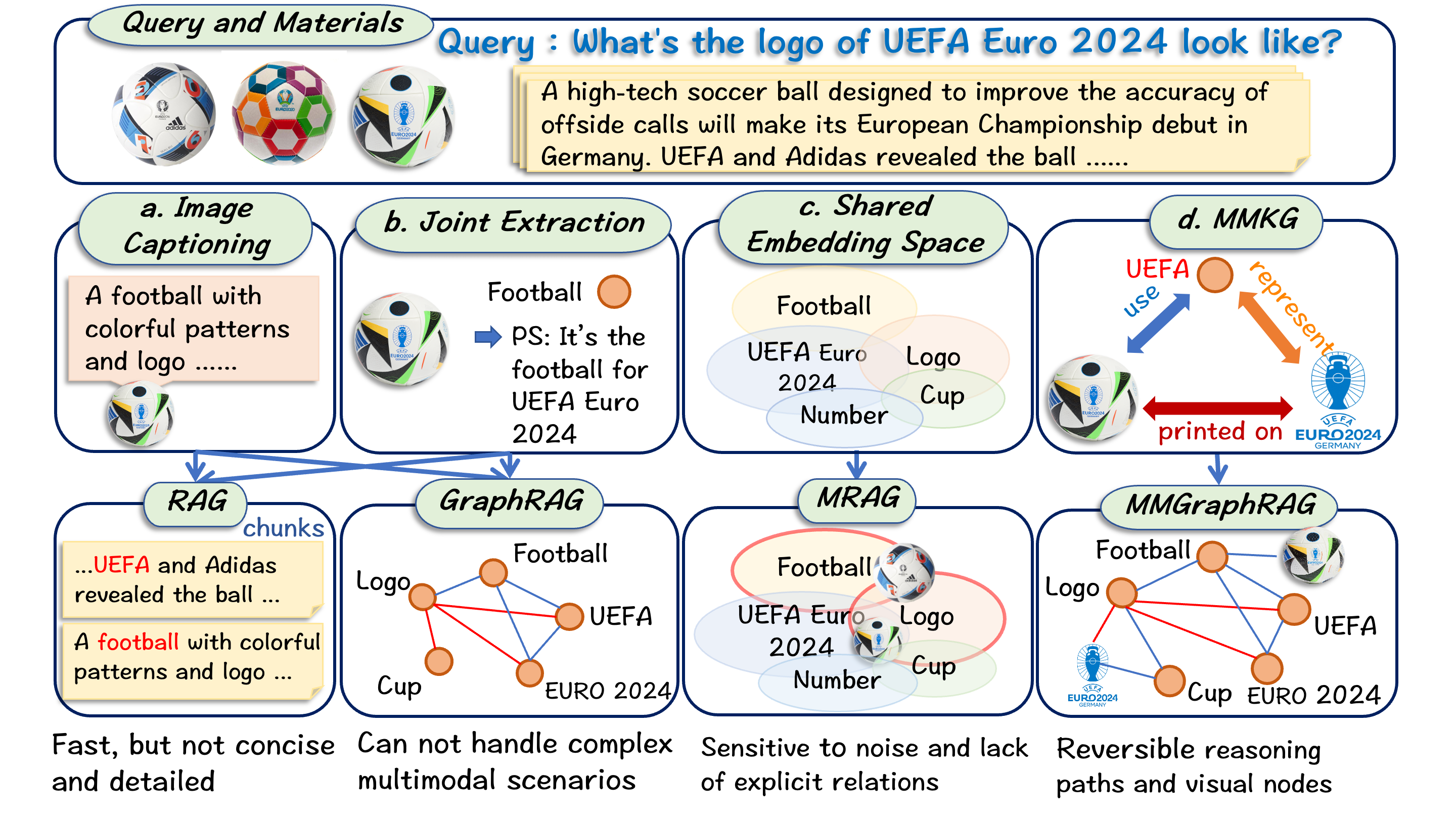}
\caption{Comparison of image-text fusion paradigms. Prior approaches either lose structural information (captioning), rely on strong supervision (joint extraction), or flatten semantics into embedding space. In contrast, MMGraphRAG represents visual and textual entities as nodes in a unified multimodal graph, enabling explicit cross-modal reasoning.}
\label{Comparison}
\end{figure}

Existing multimodal fusion methods struggle to construct such structured representations. Approaches based on image captioning linearize visual content into text, losing fine-grained structure~\cite{7298935,e26100876}. Shared embedding methods map modalities into a unified vector space but obscure explicit relations~\cite{DBLP:journals/corr/abs-2410-10594,faysse2025colpali,DBLP:journals/corr/abs-2411-04952,ling2025mmkbragmultimodalknowledgebasedretrievalaugmented}. Joint extraction methods depend on task-specific annotations and lack generalizability~\cite{tsai-etal-2019-multimodal,zhou2020improving}. As illustrated in Figure~\ref{Comparison}, these approaches fail to preserve explicit entities, relations, and cross-modal evidence paths, which are essential for structured retrieval and provenance-supported generation.

To address these limitations, we propose \textbf{MMGraphRAG}, a framework for constructing document-level multimodal graph indexes that unify textual and visual evidence. Our approach represents visual content as scene graphs and integrates them with text-derived entity-relation graphs through a cross-modal entity linking (CMEL) module, \textbf{SpecLink}. By modeling both semantic similarity and graph structure via spectral clustering, SpecLink aligns visual and textual entities and turns separated modality-specific graphs into a unified retrieval structure. This design allows downstream generation to operate over explicit multimodal entities, relations, and evidence links, improving both answer completeness and provenance visibility.

A central challenge in multimodal knowledge graph (MMKG) construction is the CMEL task, which aims to establish semantic correspondence between entities across modalities~\cite{DBLP:journals/corr/abs-2305-14725}. Despite its importance, this problem remains underexplored and lacks standardized benchmarks. To fill this gap, we introduce the \textbf{\textsc{CMEL} dataset}, designed for fine-grained cross-modal entity alignment in realistic multimodal scenarios. Combined with SpecLink, this dataset provides a foundation for systematic evaluation of MMKG construction.

We evaluate MMGraphRAG on multimodal Document Question Answering (DocQA) benchmarks. Results show consistent improvements over existing methods, particularly on questions requiring multimodal evidence and on unanswerable queries. These gains highlight the importance of explicit multimodal graph construction for reliable evidence organization.

Our main contributions are as follows:
\begin{itemize}
	\item We propose \textbf{MMGraphRAG}, a framework that constructs a document-level multimodal graph index by representing visual content as scene graphs and integrating it with text-derived entity-relation graphs.
	\item We introduce \textbf{SpecLink}, a spectral-clustering-based CMEL module that aligns visual and textual entities by jointly considering semantic similarity and graph structure.
	\item We construct the \textbf{CMEL dataset}, a benchmark for fine-grained cross-modal entity alignment in realistic multimodal documents, enabling systematic evaluation of the graph-construction stage.
\end{itemize}

\section{Related Work}

\subsection{Graph-Based RAG for Multimodal Reasoning}

The integration of external knowledge through RAG has become a standard paradigm for grounding LLMs. To support complex multi-hop reasoning, recent work has moved beyond naive vector retrieval toward graph-structured representations. GraphRAG~\cite{edge2025localglobalgraphrag} introduces entity-centric knowledge graphs and community-level summaries to capture relational context, while subsequent approaches further enhance retrieval, reasoning, and incremental graph access through structured graph organization~\cite{guo2024lightrag,li2025structrag,ma2025thinkongraph,li2025simple}. These methods highlight the importance of explicit relational structures for evidence organization. However, they remain largely text-centric and do not explicitly model multimodal knowledge.

\begin{figure*}[t]
\centering
\includegraphics[width=\textwidth]{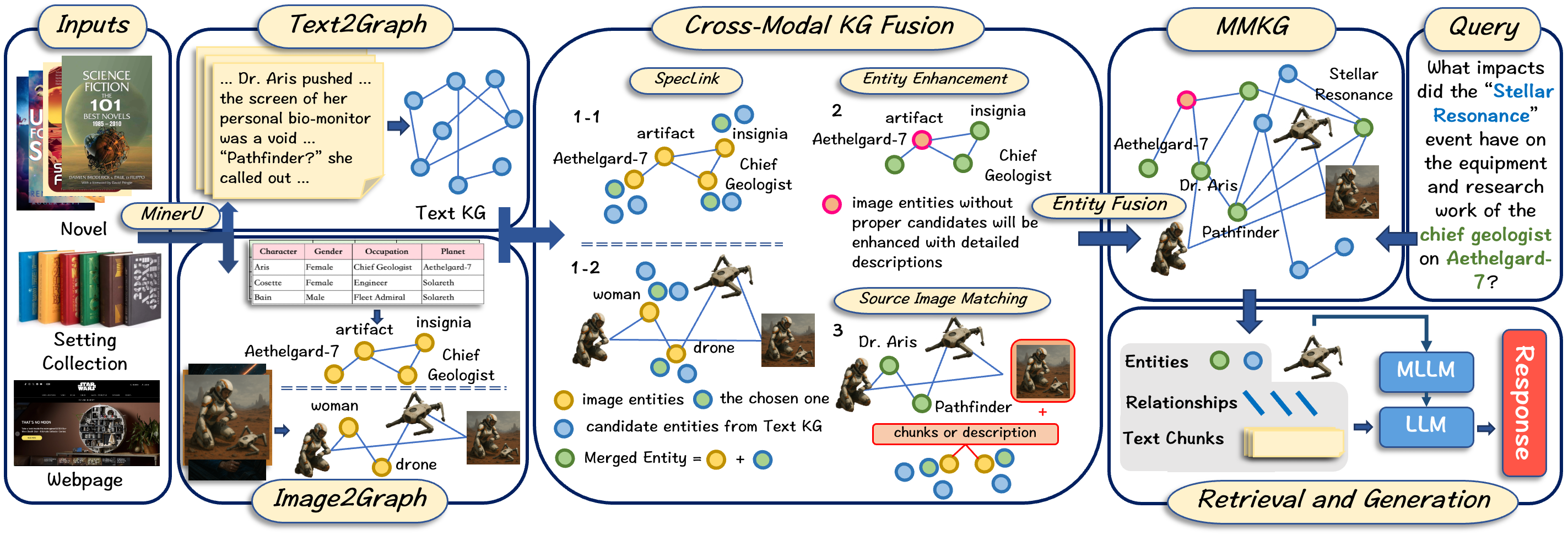}
\caption{MMGraphRAG Framework Overview. The framework constructs modality-specific entity-relation graphs via Text2Graph and Image2Graph, and integrates them through cross-modal entity linking (SpecLink) to form a unified multimodal graph index. This graph provides structured context for multimodal retrieval and generation.}
\label{framework}
\end{figure*}

It is useful to distinguish graph \emph{construction} from graph \emph{retrieval and reasoning}. Many graph-based LLM methods assume an existing graph or build auxiliary text-only graphs, then focus on how to traverse, retrieve from, or reason over that structure. By contrast, MMGraphRAG focuses on the index-construction stage: it builds a persistent multimodal graph directly from raw image-text documents without relying on a pre-existing symbolic KG. Retrieval-oriented graph reasoning methods are therefore complementary to our framework and can, in principle, operate on the multimodal graph produced by MMGraphRAG.

In multimodal settings, existing approaches struggle to preserve structured visual evidence. Methods such as HM-RAG~\cite{liu2025hmraghierarchicalmultiagentmultimodal} incorporate heterogeneous data but convert visual content into textual descriptions, resulting in the loss of fine-grained spatial and relational information. Other approaches construct query-specific multimodal graphs at inference time~\cite{bu2025query}, improving flexibility but lacking persistent and reusable graph indexes. As a result, these methods do not provide a unified representation that supports consistent cross-modal evidence retrieval. This reveals a key limitation: current RAG-based systems lack the ability to construct explicit multimodal entity-relation graphs that preserve both structural and cross-modal information.

\subsection{Multimodal Knowledge Graphs}

The term KG is used across several related but distinct traditions. In the Semantic-Web sense, a KG is commonly understood as an ontology-grounded graph, often expressed through the Resource Description Framework (RDF) or property-graph formalisms, with controlled vocabularies and possible entailment or consistency reasoning~\cite{ehrlinger2016towards}. Globally curated resources such as DBpedia, Wikidata, and Freebase exemplify this formal, symbolic view. MMKGs extend this notion by associating entities or relations with non-symbolic modalities, especially images. Recent surveys distinguish attribute-style MMKGs, which attach images or visual features to symbolic entities, from object-centric MMKGs, which promote visual objects to first-class graph nodes~\cite{zhu2024multimodal,liang2024survey,peng2023multimodal}. Representative systems such as IMGpedia and MMKG enrich existing KG backbones with visual descriptors, cross-KG signals, or multimodal embeddings~\cite{ferrada2017imgpedia,liu2019mmkg,thoma2017towards}.

Prior MMKG work motivates explicit multimodal structure, but it still leaves open the problem addressed here: constructing a persistent, document-level visual-textual entity-relation index directly from raw image-text documents without a curated symbolic backbone or manual ontology design.

\subsection{Cross-Modal Entity Linking}

The integration of multimodal information in knowledge systems has evolved from implicit fusion toward explicit semantic alignment. Early approaches rely on shared embedding spaces to connect images and text~\cite{radford2021learning,faysse2025colpali,ling2025mmkbragmultimodalknowledgebasedretrievalaugmented}. While effective for retrieval, these methods flatten multimodal semantics into dense vectors, obscuring explicit entities and relations. This limitation reduces interpretability and can lead to unreliable reasoning, especially in complex scenarios. For example, M3DocRAG~\cite{DBLP:journals/corr/abs-2411-04952} shows poor performance on unanswerable queries, highlighting the fragility of reasoning over implicit representations.

To address these limitations, multimodal entity linking (MEL) methods introduce explicit associations between modalities. Traditional entity linking focuses on aligning textual mentions with knowledge base entries~\cite{6823700,9560019}, while MEL incorporates visual signals to improve disambiguation~\cite{10.1145/3474085.3475400,10.1145/3627673.3679793}. However, most MEL approaches treat visual information as auxiliary attributes rather than first-class entities, limiting their ability to model rich cross-modal relationships.

The CMEL paradigm advances this direction by treating visual content as explicit entities and aligning them with textual counterparts to construct multimodal graphs~\cite{DBLP:journals/corr/abs-2305-14725}. In this paper, CMEL refers to document-level alignment between visual entities and text entities rather than linking mentions to an external knowledge base (KB). This enables structured cross-modal evidence organization over a unified entity-relation representation. Nevertheless, current CMEL research remains limited by the lack of realistic benchmarks and robust evaluation protocols. Existing datasets often rely on synthetic environments or simplified scenarios~\cite{alonso2025visionlanguagemodelsstrugglealign}, which do not capture the complexity of real-world multimodal data. This gap motivates the need for more generalizable alignment methods and datasets that support fine-grained, structure-preserving multimodal graph construction.
\section{Methodology}
\label{sec:methodology}

We propose \textbf{MMGraphRAG}, a framework for constructing and utilizing multimodal graph indexes to support structure-aware retrieval and generation. In this paper, the constructed MMKG is a lightweight, retrieval-oriented entity-relation graph stored in GraphML. Nodes correspond to textual or visual entities and carry free-text descriptions; edges correspond to lexical relations with relation descriptions and LLM-assigned importance weights. The graph does not rely on RDF typing, formal ontologies, or entailment rules. Instead, it serves as a persistent and reusable index that preserves textual entities, visual entities, cross-modal links, and evidence descriptions for downstream multimodal document question answering.

\subsection{MMGraphRAG Framework}

As illustrated in Figure~\ref{framework}, MMGraphRAG consists of three stages: \textit{Indexing}, \textit{Retrieval}, and \textit{Generation}. The framework constructs a persistent multimodal graph index that serves as an independent evidence store, enabling consistent reuse across queries.

\textbf{Indexing.} The indexing stage transforms raw multimodal inputs into a structured MMKG. It includes three components:
\begin{itemize}
    \item \textbf{Preprocessing:} Documents are parsed (e.g., using \textit{MinerU}~\cite{wang2024mineru}) to separate textual and visual content and normalize them for downstream processing.
    \item \textbf{Single-Modal Graph Construction:} Text is converted into a text-based entity-relation graph via chunking and entity extraction, while images are mapped to graph structures through scene graph construction.
    \item \textbf{Cross-Modal Fusion:} The modality-specific graphs are unified via cross-modal entity linking using the proposed SpecLink method, forming a coherent multimodal graph index.
\end{itemize}

The text-side component follows the entity and relation extraction paradigm of GraphRAG~\cite{edge2025localglobalgraphrag}: an LLM extracts typed entities, relation descriptions, and supporting text units from each chunk, followed by same-name merging. We re-implement this process so that image-derived entities and relations can enter the same graph store as text-derived nodes. Entity types use a lightweight default type set, and relations carry LLM-assigned importance weights that are later used by SpecLink.

\textbf{Retrieval and Generation.} 
Given a query, retrieval is performed as bounded local subgraph selection over the constructed multimodal graph, returning a set of relevant entities and relations that may span both textual and visual modalities. Compared to text-only GraphRAG, this process operates on a multimodal graph and thus incorporates visual entities into the retrieved context.

Retrieval proceeds in three steps. First, the query is embedded and matched against the entity vector index, and the top-$k$ most relevant entities are selected as seed nodes. Second, the system expands one hop from these seeds along incident relations and collects supporting text units, forming a local subgraph. Third, the retrieved context is truncated by token budgets for relations and supporting text units, while the number of attached image entities is also capped. Thus the retrieved subgraph is bounded by seed count, graph expansion depth, and context budgets rather than by the full graph size.

The retrieved subgraph is used as structured input for generation. The LLM performs textual reasoning and generation, while the Multimodal Large Language Model (MLLM) processes associated visual evidence when necessary.

\subsection{Image2Graph}

To enable MMKG construction, visual content must be transformed into structured, entity-centric representations. Scene graphs provide a natural abstraction, but traditional methods often miss fine-grained semantics and implicit relations, leading to incomplete structures and biased downstream reasoning~\cite{liu2025hmraghierarchicalmultiagentmultimodal,liu2025aligningvisionlanguagetextfree}.

We adopt MLLM-based extraction to construct richer scene graphs. Leveraging their multimodal understanding, MLLMs can identify entities and infer both explicit spatial relations (e.g., \textit{girl holding camera}) and implicit semantic relations (e.g., social or contextual associations). They also generate more informative entity descriptions, improving semantic granularity without relying on large-scale annotated datasets~\cite{Lin_2020_CVPR,Suhail_2021_CVPR,Tang_2020_CVPR,Xu_2017_CVPR,Zellers_2018_CVPR,Zheng_2023_CVPR}.

\begin{figure}[ht]
\centering
\includegraphics[width=\linewidth]{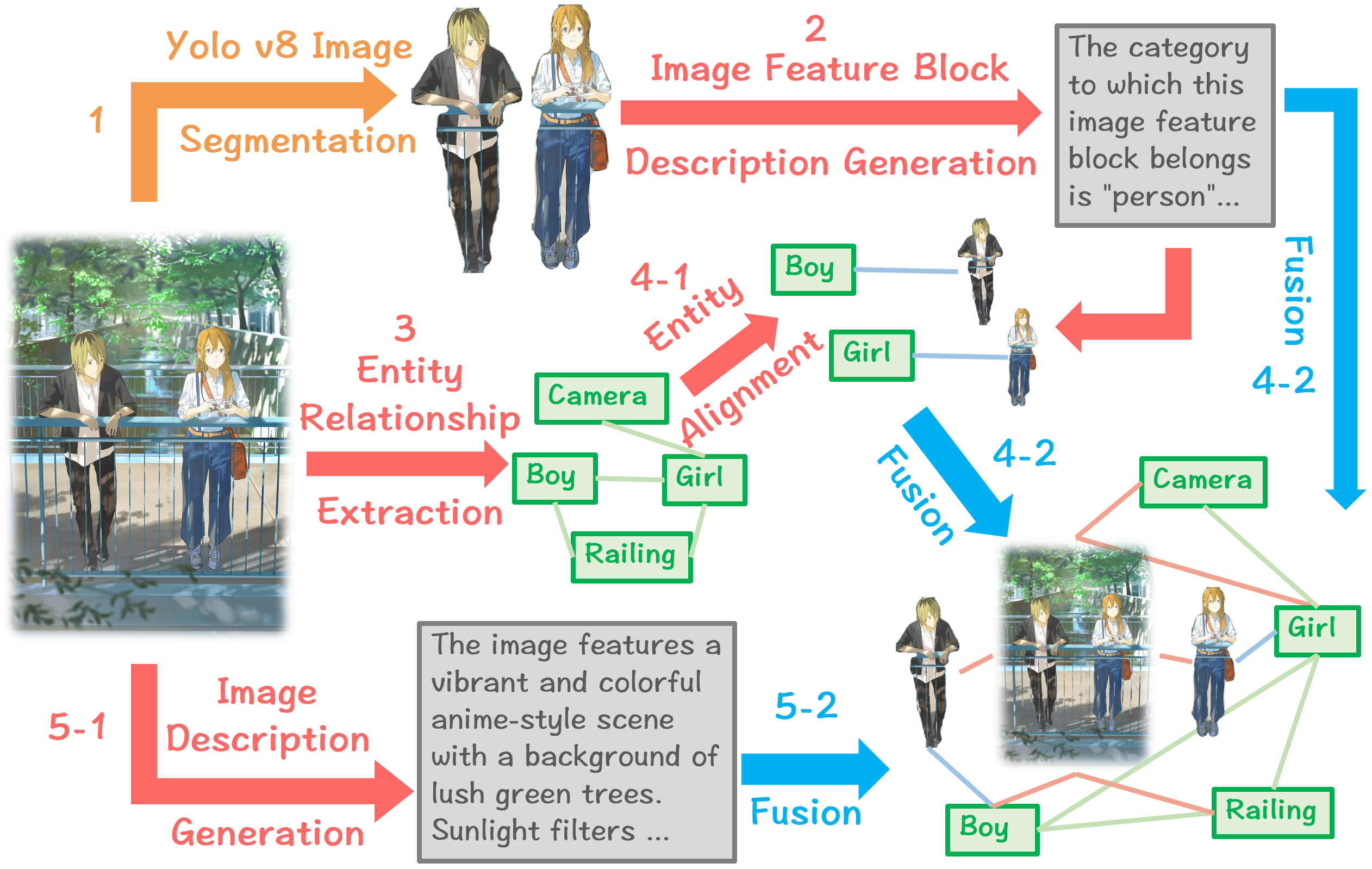}
\caption{An Example of the Img2Graph Module in Action}
\label{image2graph}
\end{figure}

The \textit{Img2Graph} module converts an image into an entity-relation graph through the following steps: 
(1) You Only Look Once (YOLO) semantic segmentation~\cite{yolov8} to obtain region-level feature blocks; 
(2) MLLM-based description of each region; 
(3) extraction of entities and relations; 
(4) alignment between regions and entities; 
and (5) construction of a global image entity linked to local entities. An example is shown in Figure~\ref{image2graph}.

The resulting graph represents visual content as explicit entities and relations, forming a structured representation that preserves spatial layout and entity descriptions. YOLO segmentation enriches entity granularity and helps retrieval return more precise visual subgraphs. A missed object does not break graph construction because the MLLM still performs regional and global image extraction; instead, false negatives mainly reduce the granularity of the resulting image graph. This structured representation enables consistent integration with textual graphs and supports multimodal evidence retrieval.

\subsection{Cross-modal Fusion Module}

The objective of cross-modal fusion is to construct a unified multimodal graph by aligning and integrating entities from image-based and text-based graphs. This process establishes explicit entity-level correspondences across modalities and enables a coherent graph representation.

\subsubsection{Fine-Grained Entity Alignment Between KGs}

Cross-modal fusion is formulated as a CMEL task, which aligns visual entities with their textual counterparts.

\paragraph{CMEL Task Definition}

Define image set as \( I = \{ I_1, I_2, \dots, I_N \} \), where each image \( I_i \) contains a set of extracted entities denoted by: \(E(I_i) = \{ e_1^{(I_i)}, e_2^{(I_i)}, \dots, e_K^{(I_i)} \}.\)

Similarly, define the text set as \( T = \{ T_1, T_2, \dots, T_M \} \), where each text chunk \( T_j \) contains a set of extracted entities: \(E(T_j) = \{ e_1^{(T_j)}, e_2^{(T_j)}, \dots, e_L^{(T_j)} \}.\)

The goal of CMEL is to identify pairs of entities from images and text that refer to the same real-world concept within the document. In other words, for each visual entity \( e_k^{(I_i)} \), we aim to align it with the most semantically relevant textual entity \( e_l^{(T_j)} \). Since the number of textual entities is generally larger than that of visual entities, the task is decomposed into two stages: (1) generating a set of candidate textual entities for each visual entity, and (2) selecting the best-aligned textual entity from this set.

Formally, for each visual entity \( e_k^{(I_i)} \), define the candidate set as:
\[
C(e_k^{(I_i)}) \subseteq \bigcup_{j-1}^{j+1} E(T_j),
\]
where \( C(e_k^{(I_i)}) \) contains the most relevant textual entities selected from the textual entity pool of the context.

The final alignment is determined by maximizing a similarity function \( f \), such that:
\[
\mathcal{A}(e_k^{(I_i)}) = \arg \max_{e \in C(e_k^{(I_i)})} f(e_k^{(I_i)}, e).
\]

Once aligned, the linked entity pairs are passed to an LLM-based module to ensure they share a unified representation in the KG.

\paragraph{SpecLink:A Spectral Clustering\mbox{-}Based Candidate Generation}

To improve the efficiency and robustness of candidate entities generation, we propose SpecLink, a spectral clustering\mbox{-}based optimization strategy. Existing methods fall into two categories: (1) \textit{distance-based clustering}, such as KMeans \cite{kodinariya2013review,LIKAS2003451,9072123} and Density-Based Spatial Clustering of Applications with Noise (DBSCAN) \cite{9356727,6814687}, which depends on semantic similarity but ignores graph structure, and (2) \textit{graph-based clustering}, such as PageRank \cite{ATABRIZI20135772} and Leiden \cite{traag2019louvain}, which captures structural relations but suffers in sparse graphs. To address both aspects, we design SpecLink tailored for CMEL.

Specifically, we redesign the weighted adjacency matrix \( \mathbf{A} \) and the degree matrix \( \mathbf{D} \) to capture both semantic and structural information between entities.

The adjacency matrix \( \mathbf{A} \) is constructed to reflect the similarity between nodes as well as the importance of their relations. It is defined as:

\begin{equation} 
\mathbf{A}_{pq} = sim(\mathbf{v}_p, \mathbf{v}_q) \cdot weight(r_{pq}) 
\end{equation}

where \( \mathbf{v}_p \) is the embedding vector of entity \( e_p \), and $sim(\cdot)$ denotes cosine similarity. The term \( r_{pq} \) represents the relation between \( e_p \) and \( e_q \) in the KG, and $weight(r_{pq})$ is a scalar reflecting the importance of the relation assessed by LLMs. If no relation exists between two entities, we set $weight(r_{pq}) = 1$.

The degree matrix \( \mathbf{D} \) is a diagonal matrix, where each diagonal entry \( \mathbf{D}_{pp} \) indicates the connectivity strength of node \( p \), computed as:
\[
\mathbf{D}_{pp} = \sum_q \mathbf{A}_{pq}.
\]
Intuitively, each diagonal value in \( D \) represents the total weighted similarity between node \( p \) and all other nodes.

Following the standard spectral clustering procedure, we construct the Laplacian matrix and perform eigen-decomposition. We then form the matrix \( \mathbf{Q} = [\mathbf{u}_1, \dots, \mathbf{u}_m]\) using the smallest \( m \) eigenvectors, where \( m \) depends on the number of textual entities in context \cite{jia2014latest}.

Clustering is performed on the row space of \( \mathbf{Q} \) using DBSCAN to obtain cluster partitions:
\[
\text{Cluster}(\mathbf{Q}) = \{C_1, C_2, \dots, C_n\}.
\]
In implementation, the candidate text entity pool is collected from the neighboring text chunks around the image chunk, with boundary clipping at the beginning and end of a document. If this pool contains \(n\) text entities, the spectral embedding dimension \(m\) is set as a function of \(n\). Entity descriptions are embedded with the same sentence encoder, and DBSCAN is applied to the spectral features. The image entity is then assigned to a cluster by nearest-neighbor matching over text-entity embeddings with cosine distance, and all text entities in the assigned cluster become its candidate set. The neighborhood size and clustering hyperparameters used in experiments are reported in Section~\ref{sec:cmel-experiments}.

For each image entity \( e_k^{(I_i)} \), we select the most relevant cluster based on the cosine similarity between its embedding \( \mathbf{v}_k^{(I_i)} \) and the cluster members. The candidate entity set is then defined as:
\[
C(e_k^{(I_i)}) = \bigcup_{C_n} \{ e_p \mid e_p \in C_n \}.
\]

Finally, we perform entity alignment using LLM-based inference, which has demonstrated high accuracy and adaptability in complex alignment scenarios \cite{10.1145/3627673.3679793,liu2024onenet}. In SpecLink, the LLM is used only for the final in-candidate alignment decision, while candidate generation is handled by the spectral procedure above. The prompt includes:
\begin{itemize}
    \item the name and description of the visual entity,
    \item descriptions of candidate entities from the selected cluster, and
    \item a fixed set of alignment examples.
\end{itemize}
The output is adopted as the final alignment result.

\subsubsection{Entity Enrichment and Fusion}

Visual entities not aligned during CMEL are enriched using contextual information from the text to improve semantic completeness. 

For example, a text may describe an event such as \textit{``HURRICANE SLAMS COASTAL REGIONS IN WEST FLORIDA''}, while the corresponding image contains an entity such as \textit{a flooded neighborhood} without explicit alignment. By incorporating contextual cues (e.g., \textit{Hurricane Ian}, \textit{Florida}), the visual entity can be refined into a more informative representation, such as \textit{a neighborhood in Florida flooded by Hurricane Ian}. 

In addition, each image is associated with a global entity representing its overall semantics. This entity is aligned with relevant textual entities when possible, or introduced as a new node otherwise.

For aligned entities, we perform fusion to ensure consistent representation within the MMKG, resulting in a unified graph where entities are shared across modalities.

\subsubsection{Graph Construction}

The above process is applied iteratively to all images and associated texts, yielding a unified MMKG that integrates multimodal entities and relations.

\section{Experiments}

This section evaluates MMGraphRAG from three perspectives: SpecLink on CMEL, end-to-end DocQA performance on DocBench and MMLongBench, and ablations on cross-modal fusion and explicit graph representation.

\subsection{CMEL Experiments}
\label{sec:cmel-experiments}

To evaluate SpecLink, we construct CMEL, a benchmark for fine-grained multi-entity alignment in complex multimodal scenarios. Compared with existing benchmarks such as MATE~\cite{alonso2025visionlanguagemodelsstrugglealign}, CMEL contains greater entity diversity and relation complexity, and supports extensible sample construction through a semi-automated pipeline. Additional construction details, examples, and reliability validation are provided in the Appendix.

The CMEL dataset comprises documents from three distinct domains---news, academia, and novels---ensuring broad domain diversity and practical applicability. Each sample includes (i) a text-based KG built from text chunks, (ii) an image-based KG derived from per-image scene graphs, and (iii) the original PDF-format document. In total, CMEL provides 1,114 alignment instances: 87 drawn from news articles, 475 from academic papers, and 552 from novels. Gold alignments are constructed through a semi-automatic entity fusion pipeline followed by manual verification. A random manual inspection of 20\% of the data reports 97.7\% coverage of entities requiring fusion and 99.0\% correctness of the merged entities, indicating that the evaluation labels are highly reliable.

To comprehensively assess performance, we adopt both micro-accuracy and macro-accuracy as evaluation metrics. Micro-accuracy is computed on a per-entity basis, reflecting the overall prediction correctness and serving as an indicator of global performance. Meanwhile, macro-accuracy calculates the average accuracy per document, mitigating evaluation bias caused by imbalanced entity distributions across documents and better highlighting performance of different methods across diverse domains.

\subsubsection{Experimental Setup and Results}

We conduct a series of comparative experiments based on the CMEL dataset. The experiments cover three categories of unsupervised approaches: embedding-based methods (Emb), LLM-based methods (LLM), and our proposed SpecLink method (Spec). We further compare against multiple mainstream clustering algorithms to provide comprehensive baselines.

For SpecLink, candidate text entities are gathered from a three-chunk neighborhood around each image chunk, i.e., the previous, current, and next chunks after boundary clipping. If this pool contains $n$ text entities, the spectral embedding dimension is set to $m=\max(2,\lceil\sqrt{n}\rceil)$. DBSCAN is applied with $\epsilon=0.5$ and $\text{min\_samples}=\max(1,\lceil n/10\rceil)$, and the image entity is assigned to a cluster by nearest-neighbor matching with cosine distance.

The embedding-based method encodes visual and textual entities with a pretrained embedding model and selects textual candidates by cosine similarity. Its decision threshold is selected by sweeping on a held-out development split, and the same tuning protocol is applied across baseline configurations.

The LLM-based method directly generates candidate sets from a visual entity, its surrounding context, and a pool of textual entities.

For clustering\mbox{-}based baselines, we include DBSCAN (DB), KMeans (KM), PageRank (PR), and Leiden (Lei).

Entity alignment within candidate sets is uniformly conducted via LLM-based reasoning. All CMEL methods share the same candidate scope, image-entity extraction procedure, and final LLM alignment prompt; they differ only in candidate generation. This design isolates the effect of candidate generation from the final alignment judge.

To ensure the robustness and generalizability of our results, evaluations are conducted on different models. Due to space limitations, we report only the best-performing configurations for each method category in the results table. Specifically, the experiments utilize the embedding model stella-en-1.5B-v5 \cite{stella2024}, the LLM Qwen2.5-72B-Instruct \cite{qwen2.5}, and the MLLM InternVL2.5-38B-MPO \cite{internvl2_5_mpo}. The reported results are the best outcomes from three runs with different random seeds.

\begin{table}[t]
\centering
\small
\begin{tabular}{@{}ccccc@{}}
\toprule
\multirow{2}{*}{\textbf{Meth.}} & \multicolumn{3}{c}{\textbf{micro/macro Acc.}} & \multirow{2}{*}{\textbf{Overall.}} \\
\cmidrule(lr){2-4}
& News & Aca. & Nov. & \\
\midrule
Emb   & 10.8/8.4   & 33.1/34.5  & 9.0/7.5    & 20.0/16.8 \\
LLM   & 33.3/24.1  & 36.8/36.1  & 17.4/20.8  & 27.1/27.0 \\
\midrule
DB    & 53.8/45.9  & \underline{60.8/58.3} & \underline{29.9/34.2}  & \underline{45.2/46.1} \\
KM    & 50.5/40.6  & 60.7/57.7  & 29.6/30.5  & 45.2/43.0 \\
PR    & 51.6/44.4  & 59.7/56.8  & 29.1/35.2  & 44.1/45.5 \\
Lei   & \underline{54.8/44.7}  & 60.5/55.5  & 29.4/30.6  & 44.8/43.6 \\
\midrule
Spec  & \textbf{65.5/56.9} & \textbf{73.3/69.9} & \textbf{31.2/39.4} & \textbf{51.8/59.2} \\
\bottomrule
\end{tabular}
\caption{Micro/macro accuracy (\%) on CMEL dataset.}
\label{CMELresults}
\end{table}

The results are shown in Table~\ref{CMELresults}. Overall, clustering\mbox{-}based methods significantly outperform the embedding-based and LLM-based methods in the CMEL task. Compared to other clustering\mbox{-}based methods, SpecLink achieves the best performance, improving micro-accuracy by about 15\% and macro-accuracy by around 30\%, clearly demonstrating its effectiveness in CMEL task.

\subsection{Multimodal DocQA Experiments}
\label{sec:docqa-experiments}

We choose DocQA as the primary evaluation task because it tests multimodal information integration, complex reasoning, and domain adaptability over long documents with diverse formats and terminology. These characteristics make multimodal DocQA a challenging benchmark for evaluating MMGraphRAG.

\subsubsection{Benchmarks}

DocBench contains 229 PDF documents from publicly available online resources, covering five domains: academia (Aca.), finance (Fin.), government (Gov.), laws (Law.), and news (News). It includes four types of questions: pure text questions (Txt.), multimodal questions (Mm.), metadata questions, and unanswerable questions (Una.). For our experiments, since the information is converted into KGs, we do not focus on metadata. Therefore, this category of questions is excluded from statistics. We strictly follow the official evaluation protocol of DocBench, which relies on an LLM judge to determine the correctness of model answers. The judge is provided with the question, the reference answer, the model prediction, and the supporting evidence extracted from the original document, and then returns a binary decision on whether the prediction is correct.

MMLongBench consists of 135 long PDF documents collected from seven different domains. It evaluates multimodal reasoning across various content types. The Locations dimension indicates whether the supporting evidence for an answer is contained within a single page (Sin.) or spans multiple pages (Mul.) of a document, while Formats represent different content sources. MMLongBench further includes annotations covering multiple modalities---such as text (Txt.), charts (Cha.), tables (Tab.), layout (Lay.), figures (Fig.), and unanswerable items (Una.)---to comprehensively assess document understanding. Following the original benchmark, we adopt the three-step evaluation protocol of MATHVISTA \cite{lu2023mathvista}: answer generation, answer extraction using an LLM, and score calculation via an LLM judge.

In all DocBench and MMLongBench experiments, the judge model is fixed to Llama3.1-70B-Instruct. The judge is invoked with temperature~0 and greedy decoding, and the judging prompts are kept identical for all compared methods. Accuracy is reported as the sole evaluation metric to measure the correctness of model predictions.

\subsubsection{Reliability of LLM-as-a-Judge Evaluation}

To assess the reliability of LLM-based evaluation, we conduct a human--LLM consistency study. Following prior work~\cite{DBLP:journals/corr/abs-2411-04952,li2025structrag,edge2025localglobalgraphrag,zou2024docbench,ma2025mmlongbench,peng2024graph}, we compare automatic judgments with human annotations.

We sample 50 documents from DocBench and 25 from MMLongBench, and collect human labels (correct vs.\ incorrect) for comparison with the LLM judge (Llama3.1-70B-Instruct).

On DocBench (251 questions), the agreement reaches \textbf{96.0\%} with Cohen's $\kappa=0.882$. On MMLongBench (222 questions), agreement is \textbf{94.6\%} with $\kappa=0.876$. These results indicate strong consistency between LLM and human judgments.

\begin{figure}[t]
\centering
\includegraphics[width=0.92\linewidth]{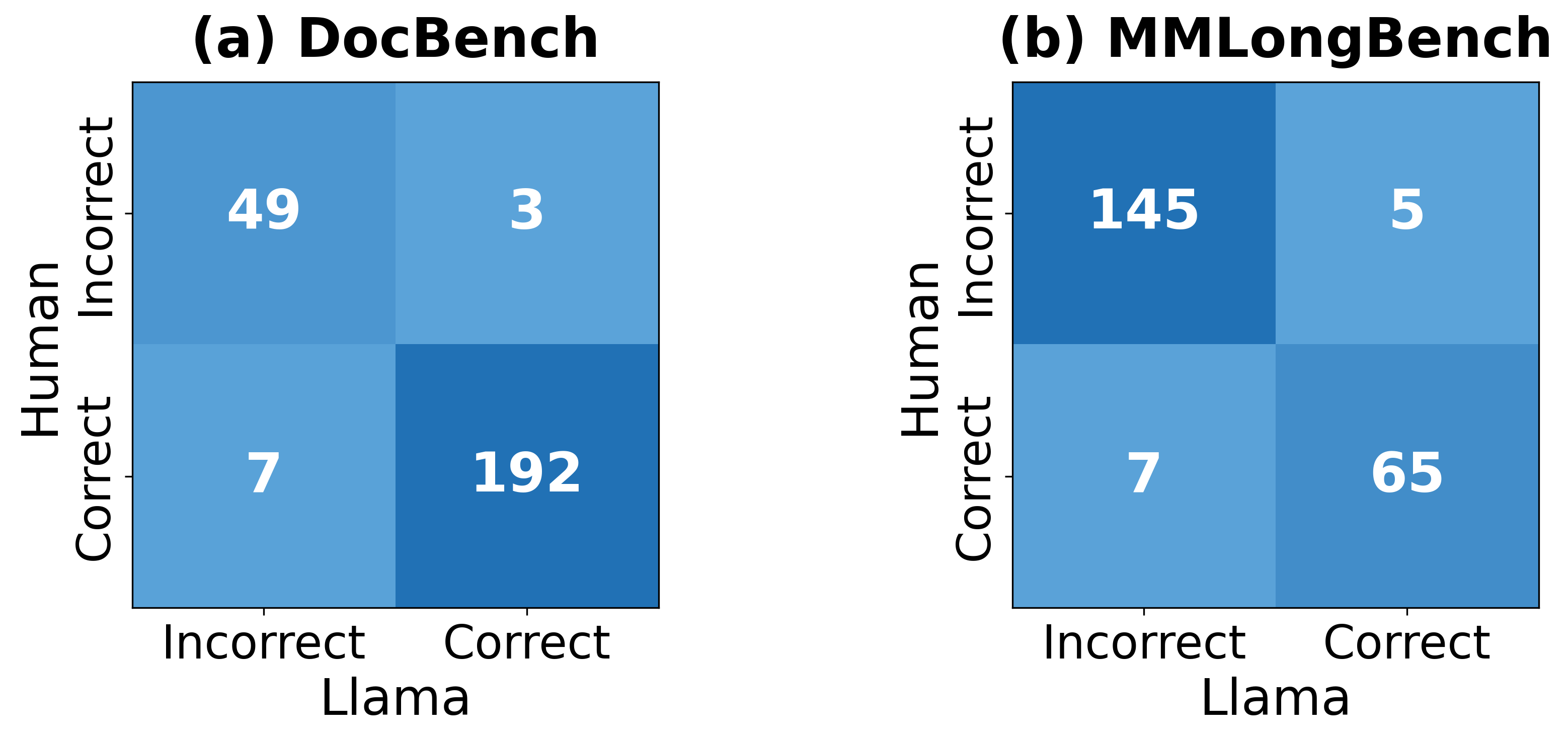}
\caption{Confusion matrices comparing human annotations and Llama3.1-70B-Instruct judgments on DocBench and MMLongBench.}
\Description{Two confusion matrices showing strong agreement between human annotations and Llama3.1-70B-Instruct judgments on DocBench and MMLongBench.}
\label{fig:judge-confusion}
\end{figure}

Figure~\ref{fig:judge-confusion} further shows that disagreements are limited to a small number of boundary cases on both benchmarks, supporting the stability of judge-based evaluation under the adopted protocol.

Overall, the high agreement suggests that LLM-based evaluation provides a reliable proxy for human assessment, with limited impact from potential biases.

\subsubsection{Comparison with Basic Methods}

In Tables~\ref{Tab.2} and~\ref{Tab.3}, L, Q, Qv, and Iv denote Llama3.1-70B-Instruct, Qwen2.5-72B-Instruct, Qwen2-VL-72B, and InternVL2.5-38B-MPO; NR/GR/MGR denote NaiveRAG/GraphRAG/MMGraphRAG.

\begin{table}[ht]
    \centering
    \small
    \setlength{\tabcolsep}{2pt}
    \begin{tabular}{@{}cccccccccc@{}}
        \toprule
         \multirow{2}{*}{\textbf{Method}} &
         \multicolumn{5}{c}{\textbf{Domains}} &
         \multicolumn{3}{c}{\textbf{Modalities}} &
         \multirow{2}{*}{\textbf{\makecell[c]{Overall \\ Acc.}}}\\
            \cmidrule(lr){2-6}
            \cmidrule(lr){7-9}
                & Aca. & Fin. & Gov. & Laws & News & Text. & Multi. & Una.\\        
        \midrule
        LLM(L) & 43.9 & 13.5 & 53.4 & 44.5 & \underline{79.7} & 52.9 & 18.8 & \textbf{81.5} & 44.7 \\
        LLM(Q) & 41.3 & 16.3 & 50.7 & 49.7 & 77.3 & 53.9 & 20.1 & 75.8 & 44.8 \\
        \midrule
        MLLM(Qv) & 17.5 & 14.9 & 25.0 & 34.6 & 48.8 & 34.0 & 8.4 & 40.3 & 25.4 \\
        MLLM(Iv) & 19.8 & 16.3 & 28.4 & 31.4 & 46.5 & 35.7 & 15.9 & 39.5 & 27.7 \\
        \midrule
        NR(L) & 43.6 & 38.2 & 66.2 & 64.9 & \textbf{80.2} & 79.9 & 32.1 & 70.2 & 61.0 \\
        NR(Q) & 43.6 & 34.4 & 62.8 & 65.4 & 75.0 & \underline{81.6} & 30.5 & 67.7 & 59.5 \\
        \midrule
        GR(L) & 40.6 & 27.1 & 56.8 & 59.7 & 75.0 & 73.5 & 24.4 & \underline{76.6} & 54.7 \\
        GR(Q) & 39.6 & 25.7 & 52.5 & 49.7 & 74.5 & 71.7 & 26.0 & 67.5 & 52.3 \\
        \midrule
        \textbf{MGR(L-Qv)} & 51.8 & 59.4 & 62.8 & 60.7 & 77.9 & 79.1 & 77.8 & 70.2 & 74.0 \\
        \textbf{MGR(Q-Qv)} & 51.8 & 62.9 & \textbf{66.9} & \underline{68.6} & 76.2 & \textbf{82.4} & 81.1 & 67.7 & 75.2 \\
        \textbf{MGR(L-Iv)} & \textbf{60.7} & \underline{64.1} & 62.6 & 64.9 & 76.2 & 80.0 & \underline{86.4} & 75.0 & \textbf{77.5} \\
        \textbf{MGR(Q-Iv)} & \underline{60.5} & \textbf{65.8} & \underline{66.5} & \textbf{70.4} & 77.1 & 81.2 & \textbf{88.7} & 71.9 & \underline{76.8} \\
        \bottomrule
        \end{tabular}
    \caption{Accuracy of different methods on DocBench across domains and modalities}
    \label{Tab.2}
\end{table}

\begin{table}[ht]
    \centering
    \small
    \setlength{\tabcolsep}{2pt}
    \begin{tabular}{@{}cccccccccc@{}}
        \toprule
         \multirow{2}{*}{\textbf{Method}} &
         \multicolumn{5}{c}{\textbf{Formats}} &
         \multicolumn{3}{c}{\textbf{Locations}} &
         \multirow{2}{*}{\textbf{\makecell[c]{Overall \\ Acc.}}} \\
            \cmidrule(lr){2-6}
            \cmidrule(lr){7-9}
                & Cha. & Tab. & Txt. & Lay. & Fig. & Sin. & Mul. & Una. & \\        
        \midrule
        LLM(L) & 16.3 & 12.7 & 32.1 & 21.9 & 17.4 & 23.7 & 20.3 & 51.6 & 28.2 \\
        LLM(Q) & 16.8 & 12.6 & \underline{33.3} & 23.5 & 16.1 & 22.5 & 20.0 & 53.2 & 27.8 \\
        \midrule
        MLLM(Qv) & 7.6 & 5.4 & 10.0 & 8.8 & 12.7 & 10.8 & 9.9 & 8.1 & 10.0 \\
        MLLM(Iv) & 8.8 & 8.1 & 10.7 & 11.8 & 11.4 & 13.3 & 7.9 & 13.9 & 11.6 \\
        \midrule
        NR(L) & 16.3 & 15.3 & 31.0 & 22.7 & 18.7 & 24.9 & 19.5 & 56.5 & 29.2 \\
        NR(Q) & 15.1 & 10.8 & 30.3 & 20.3 & 14.9 & 22.3 & 16.4 & 52.5 & 26.2 \\
        \midrule
        GR(L) & 7.6 & 6.7 & 25.1 & 15.0 & 10.6 & 16.3 & 12.3 & \textbf{78.5} & 27.2 \\
        GR(Q) & 14.0 & 11.0 & 26.1 & 12.2 & 8.5 & 18.2 & 13.2 & \underline{77.1} & 28.1 \\
        \midrule
        \textbf{MGR(L-Qv)} & 26.4 & 29.2 & 26.4 & 13.8 & 21.2 & 37.6 & 13.8 & 55.2 & 32.6 \\
        \textbf{MGR(Q-Qv)} & 26.2 & 30.0 & 29.3 & \textbf{29.1} & \underline{29.2} & \underline{38.7} & 20.1 & 51.6 & 34.8 \\
        \textbf{MGR(L-Iv)} & \underline{34.7} & \textbf{36.6} & 31.9 & 12.9 & 28.5 & 38.7 & \underline{21.9} & 59.2 & \underline{36.9} \\
        \textbf{MGR(Q-Iv)} & \textbf{34.7} & \underline{36.5} & \textbf{33.8} & \underline{28.7} & \textbf{34.6} & \textbf{39.6} & \textbf{26.7} & 55.8 & \textbf{38.8} \\
        \bottomrule
        \end{tabular}
    \caption{Accuracy of different methods on MMLongBench across formats and locations}
    \label{Tab.3}
\end{table}

To ensure fair comparisons and isolate the contribution of multimodal knowledge graph construction, we adopt a unified retrieval setting across all methods. Notably, prior multimodal GraphRAG approaches primarily extend the retrieval stage, while incorporating visual information through image captioning. Under a shared retrieval pipeline, such approaches effectively reduce to text-based GraphRAG with captions replacing images, resulting in similar underlying knowledge graph structures. Therefore, we focus on representative baselines that reflect differences in knowledge representation rather than retrieval strategies.

For all DocQA RAG variants, we fix the retrieval configuration throughout the comparison. NaiveRAG retrieves the top-$k=10$ relevant text chunks. GraphRAG and MMGraphRAG retrieve the top-$k=10$ seed entities and expand one hop along incident relations. For MMGraphRAG, the retrieved context is truncated with a 6000-token relation budget and a 4000-token supporting-text budget, and at most three image entities are attached to the context.

We evaluate multiple LLMs and MLLMs to assess generality, including Llama3.1-70B-Instruct~\cite{llama3.1-70b-instruct}, Qwen2.5-72B-Instruct~\cite{yang2024qwen2}, Qwen2-VL-72B~\cite{wang2024qwen2}, and InternVL2.5-38B-MPO~\cite{chen2024internvl}.

We consider the following baselines:

\textbf{LLM}: Images are converted to text via an MLLM, and the resulting text, together with the question, is fed into the LLM. Inputs exceeding context length are split and concatenated.

\textbf{MLLM}: Images are processed jointly with the question using a multimodal model.

\textbf{NaiveRAG}~\cite{lewis2020retrieval} (NR): Text is chunked into 500-token segments, and relevant chunks are retrieved via embedding similarity under the unified retrieval configuration.

\textbf{GraphRAG} (GR): We adopt a simplified GraphRAG without community detection, using local entity retrieval for fair comparison~\cite{guo2024lightrag,edge2025localglobalgraphrag}.

The results for the DocBench and MMLongBench datasets are presented in Table~\ref{Tab.2} and Table~\ref{Tab.3}. MMGraphRAG consistently outperforms other baselines, with the largest gains observed in domains with complex visual structures (e.g., 60.7\% vs.\ 43.6\% and 65.8\% vs.\ 38.2\% on DocBench). It also shows substantial improvements on multimodal queries (88.7\% vs.\ 32.1\%), where effective integration of visual information is required. Similar trends are observed on MMLongBench, particularly for tasks involving charts, tables, and figures. These results demonstrate that explicit MMKG construction enables more effective multimodal knowledge integration and supports reliable, structure-aware reasoning.

\subsection{Ablation Study}

To investigate the individual contributions of our key components, we conduct a series of ablation studies. We aim to answer two primary questions: 1) How crucial is our sophisticated cross-modal fusion module? 2) What are the benefits of constructing an explicit MMKG compared to Multimodal RAG (MRAG) approaches?

\subsubsection{Impact of the Cross-Modal Fusion Module.}
To validate the necessity of our fusion module, we design a simplified baseline, MMGraphRAG (w/o Fusion). In this variant, the cross-modal fusion module is replaced with a naive method that performs entity alignment based solely on embedding similarity. An image entity and a text entity are considered aligned if the cosine similarity of their embeddings exceeds a predefined threshold (0.7 in our experiments). The threshold is selected on a held-out development split of the academia subset by grid-searching over $\{0.6, 0.65, 0.7, 0.75\}$ to maximize micro-accuracy. All other components are kept identical to the full MMGraphRAG model to ensure a controlled comparison.

\begin{figure}[h]
\centering
\includegraphics[width=\linewidth]{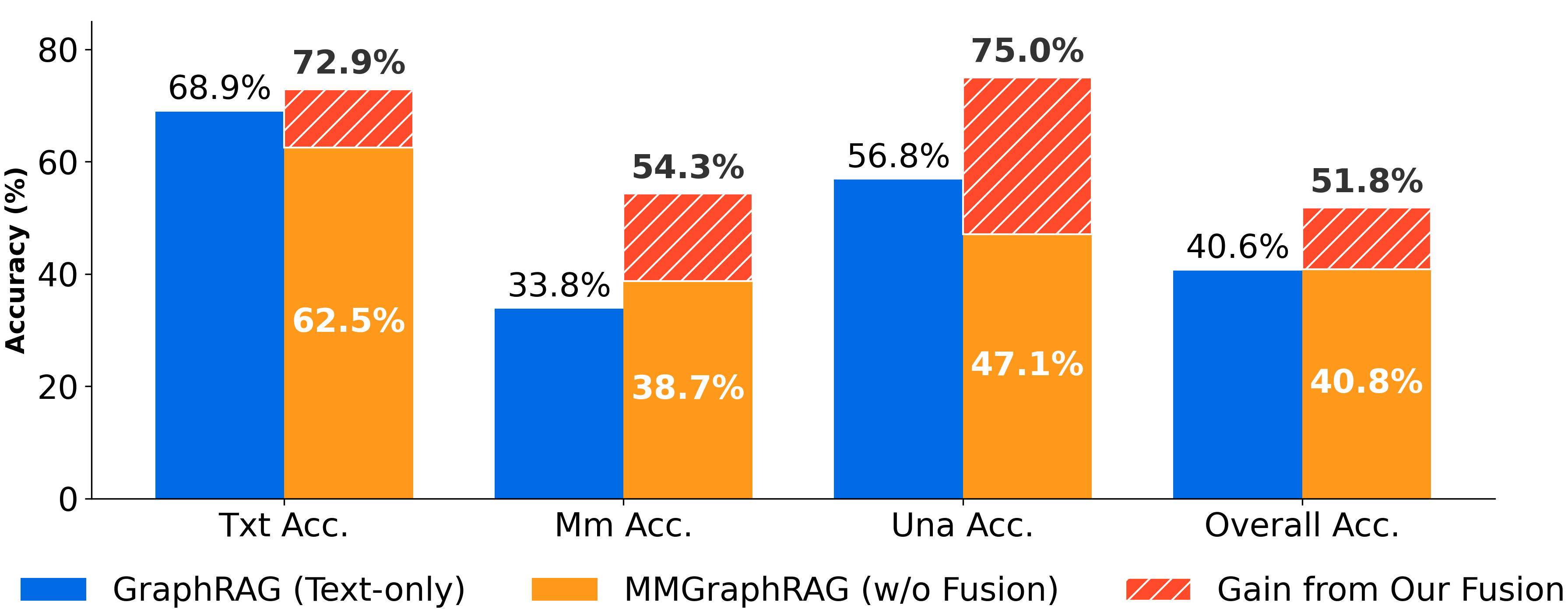}
\caption{Impact of Cross-Modal Fusion Module by Metric}
\label{ablation}
\end{figure}

We conduct this experiment on the academia (Aca) subset of DocBench. The results are presented in Figure \ref{ablation}.

The results reveal three key findings. 
(1) Naive fusion degrades performance on text-only queries, indicating that imprecise cross-modal alignment introduces noise that disrupts textual reasoning, whereas the full model improves text performance through well-aligned supplementary information. 
(2) On multimodal queries, naive fusion yields only marginal gains (+4.9\%), while MMGraphRAG achieves substantial improvements (+21.5\%), demonstrating the importance of accurate cross-modal linking. 
(3) On unanswerable questions, naive fusion significantly reduces accuracy (-9.7\%), exacerbating hallucination, whereas MMGraphRAG improves performance by 18.2\%, indicating stronger robustness. 

Overall, these results highlight that precise cross-modal alignment is critical not only for performance but also for reducing noise and improving reliability.
\subsubsection{Comparison with MRAG Methods}

To isolate the benefits of our structured MMKG, we compare MMGraphRAG against a representative MRAG framework, M3DOCRAG (M3DR) \cite{DBLP:journals/corr/abs-2411-04952}. This comparison effectively serves as an ablation of our explicit graph-building pipeline, contrasting it with approaches that typically project multimodal features into a shared embedding space for retrieval. For a fair comparison, both methods use the Qwen2.5-7B-Instruct model as the LLM and the Qwen2-VL-7B model as the MLLM, and share the same retrieval configuration (chunk size, top-$k$) and prompting strategy. The experiment is conducted on the MMLongBench dataset, and the results are shown in Table \ref{Tab.4}.

\begin{table}[h]
    \centering
    \small
    \setlength{\tabcolsep}{2pt}
    \begin{tabular}{@{}cccccccccc@{}}
        \toprule
         \multirow{2}{*}{\textbf{Method}} &
         \multicolumn{5}{c}{\textbf{Formats}} &
         \multicolumn{3}{c}{\textbf{Locations}} &
         \multirow{2}{*}{\textbf{\makecell[c]{Overall \\ Acc.}}} \\
            \cmidrule(lr){2-6}
            \cmidrule(lr){7-9}
                & Cha. & Tab. & Txt. & Lay. & Fig. & Sin. & Mul. & Una. & \\        
        \midrule
        M3DR & 18.9 & 20.1 & 30.0 & 23.5 & 20.8 & 32.4 & 14.8 & 5.8 & 21.0 \\
        MGR & \textbf{21.1} & \textbf{27.1} & 24.6 & 18.2 & 22.2 & 34.3 & 12.5 & \textbf{35.1} & \textbf{26.5} \\
        \bottomrule
    \end{tabular}
    \caption{Performance comparison between M3DOCRAG and MMGraphRAG}
    \label{Tab.4}
\end{table}

The results highlight the advantages of explicit MMKG modeling. By representing visual and textual entities and their relations, MMGraphRAG better supports cross-modal semantic alignment, leading to improved performance on structurally complex queries such as charts and tables (48.2\% vs.\ 39.0\%).

Replacing the graph structure with an MRAG-style embedding approach reveals a critical limitation in handling unanswerable questions. Embedding-based methods are more susceptible to semantically similar but irrelevant evidence, resulting in incorrect answers, whereas MMKG-based reasoning enables more reliable assessment of evidence completeness. This leads to a substantial improvement in robustness (35.1\% vs.\ 5.8\%).

Overall, these results demonstrate that explicit MMKG construction is essential for accurate and robust multimodal reasoning.
\section{Discussion and Conclusion}

This paper presents MMGraphRAG, a multimodal GraphRAG framework that constructs a lightweight entity-relation graph index from text and images and uses it for structure-aware retrieval and generation. The experiments show that explicit cross-modal graph construction improves both fine-grained entity alignment and multimodal DocQA, especially for questions involving charts, tables, figures, and incomplete evidence.

The modular design also points to a practical direction for future systems. As MLLMs become stronger, some parsing, linking, and generation steps may be integrated into fewer model calls, but an explicit multimodal graph index remains valuable because it preserves inspectable entities, relations, and evidence links. This separation between graph construction and generation is particularly useful for document-level tasks, where evidence must be reused, inspected, and compared across multiple queries rather than consumed only once by an end-to-end model. Two limitations remain. First, intra-modality deduplication is still limited, which may introduce redundancy as the graph scales. Second, downstream question answering accuracy can mask errors in the constructed graph because strong generators may compensate for imperfect retrieved context. Future work should therefore develop standalone quality metrics for multimodal graph indexes and more robust deduplication strategies.
\section{Supplemental Material Statement and Reproducibility}

To ensure reproducibility, we conduct experiments on publicly available benchmarks, including DocBench (\url{https://github.com/Anni-Zou/DocBench}) and MMLongBench (\url{https://github.com/EdinburghNLP/MMLongBench}). 

The resources for the proposed MMGraphRAG framework are hosted at \url{https://github.com/wanxueyao/MMGraphRAG}, and the CMEL dataset is hosted at \url{https://github.com/wanxueyao/CMEL-dataset}. These repositories contain the complete implementation code, dataset construction details, full prompts, and extended experimental results. Furthermore, to demonstrate the framework's efficiency and interpretability, we provide detailed time-consumption analyses for knowledge graph construction across different model scales, alongside reasoning path analyses over the MMKG.

\bibliographystyle{ACM-Reference-Format}
\bibliography{www2026}

\clearpage
\appendix
\appendixfloatnumbering
\twocolumn[
  \begin{@twocolumnfalse}
    \centering
    \vspace{2.5em} % 增加顶部间距，使其更美观
    
    % 第一行：论文原标题（加粗）
    {\huge \bfseries MMGraphRAG: Bridging Vision and Language \\ with Interpretable Multimodal Knowledge Graphs \par}
    
    \vspace{1.5em} % 两个标题之间的间距
    
    % 第二行：Supplementary Material（加粗）
    {\LARGE \bfseries Supplementary Material \par}
    
    \vspace{2.5em} % 标题与正文起始处的间距
  \end{@twocolumnfalse}
]
\section{CMEL dataset}

The CMEL dataset provides a benchmark for cross-modal entity linking, comprising 1,114 instances from three domains: news (87), academia (475), and novels (552). Each sample includes a text-based KG, an image-based KG, and the original PDF.

\subsection{A Detailed Introduction to the CMEL Dataset}

\begin{figure*}[t]
\centering
\includegraphics[width=0.86\textwidth]{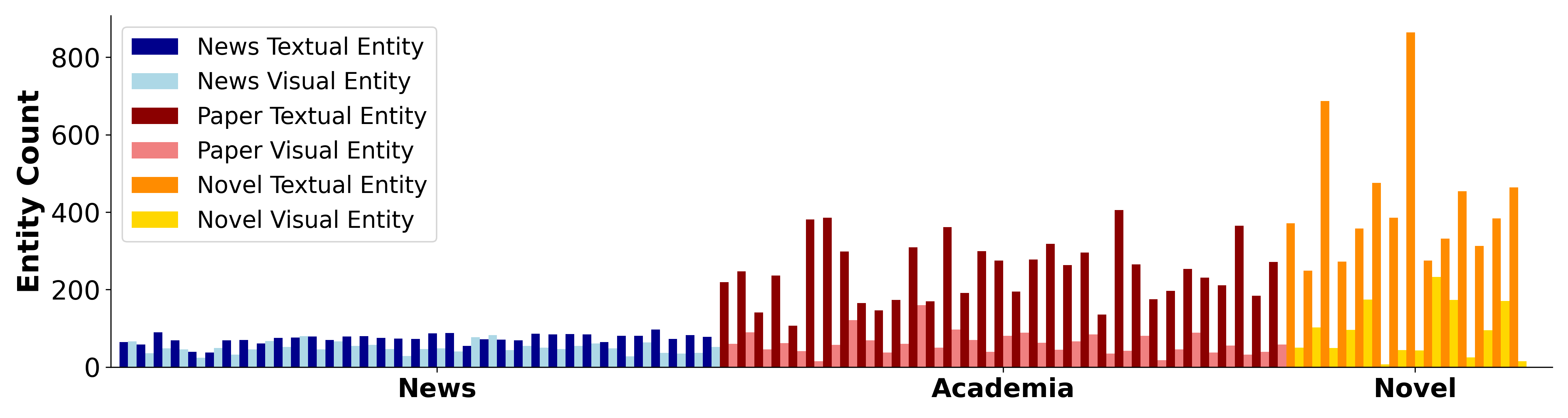}
\caption{Overall composition of the CMEL dataset, including its three domains and multimodal resources.}
\Description{A high-level overview of the CMEL dataset showing the three domains and the multimodal components in each sample, including PDF documents, text-based knowledge graphs, and image-based knowledge graphs.}
\label{Fig.s0}
\end{figure*}

Figure \ref{Fig.s0} shows the dataset composition across three domains. Each sample includes the original PDF, a text-based KG from chunks, and an image-based KG from scene graphs.

\begin{figure}[H]
\centering
\includegraphics[width=\linewidth]{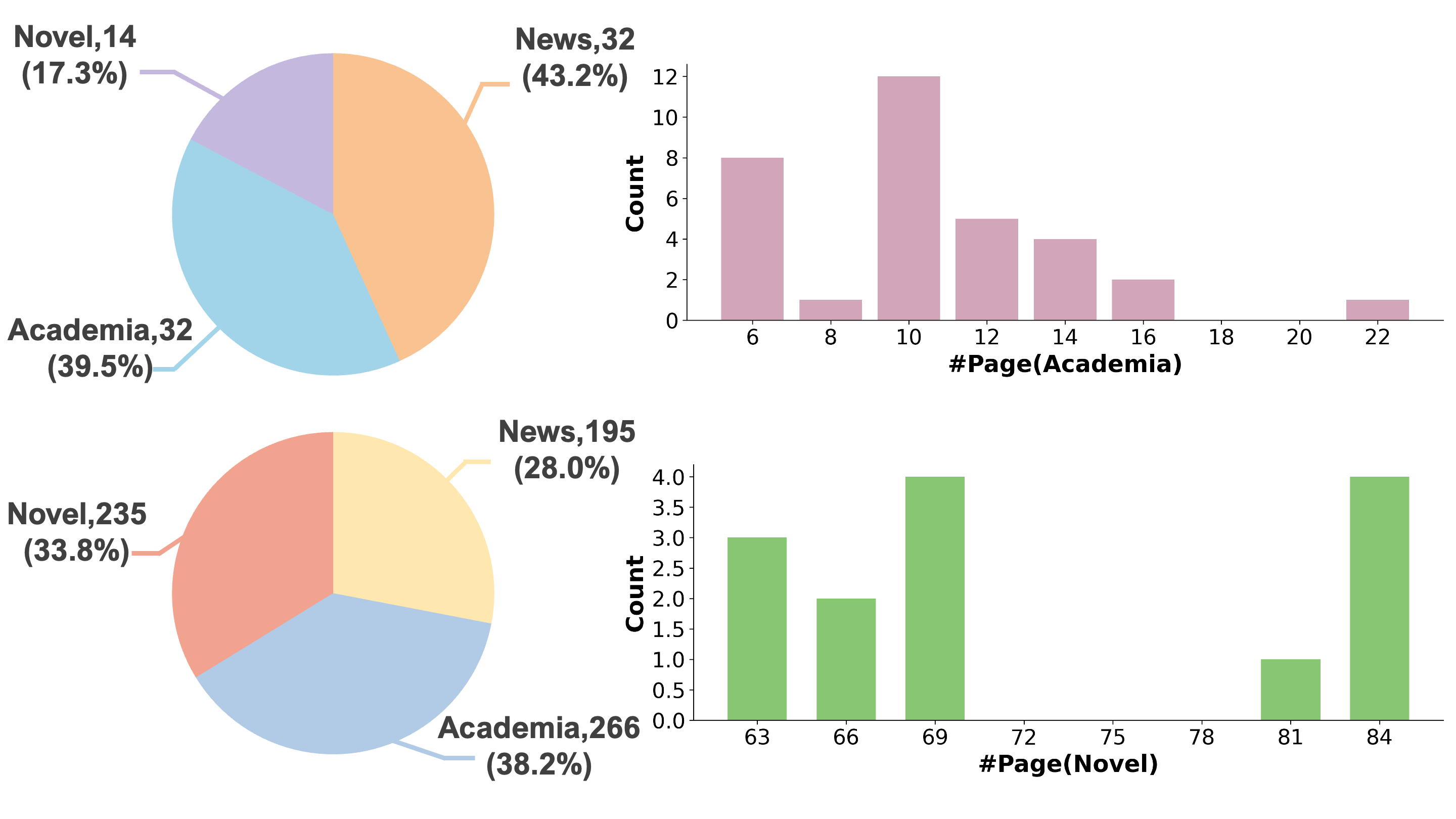}
\caption{The Distribution of CMEL dataset. In the top-left, the number and proportion of documents in each domain are shown; in the bottom-left, the number and proportion of images in each domain are displayed; in the top-right, the page distribution of academia domain documents is provided; and in the bottom-right, the page distribution of novel documents is shown. All news domain documents are one page.}
\Description{A four-panel statistical visualization of CMEL dataset distribution, including document counts by domain, image counts by domain, and page-number distributions for academia and novel documents.}
\label{Fig.s1}
\end{figure}

Figure \ref{Fig.s1} shows the distribution across domains. The dataset is constructed based on image count, ensuring approximately equal document distribution per domain.

\begin{figure}[H]
\centering
\includegraphics[width=0.76\linewidth]{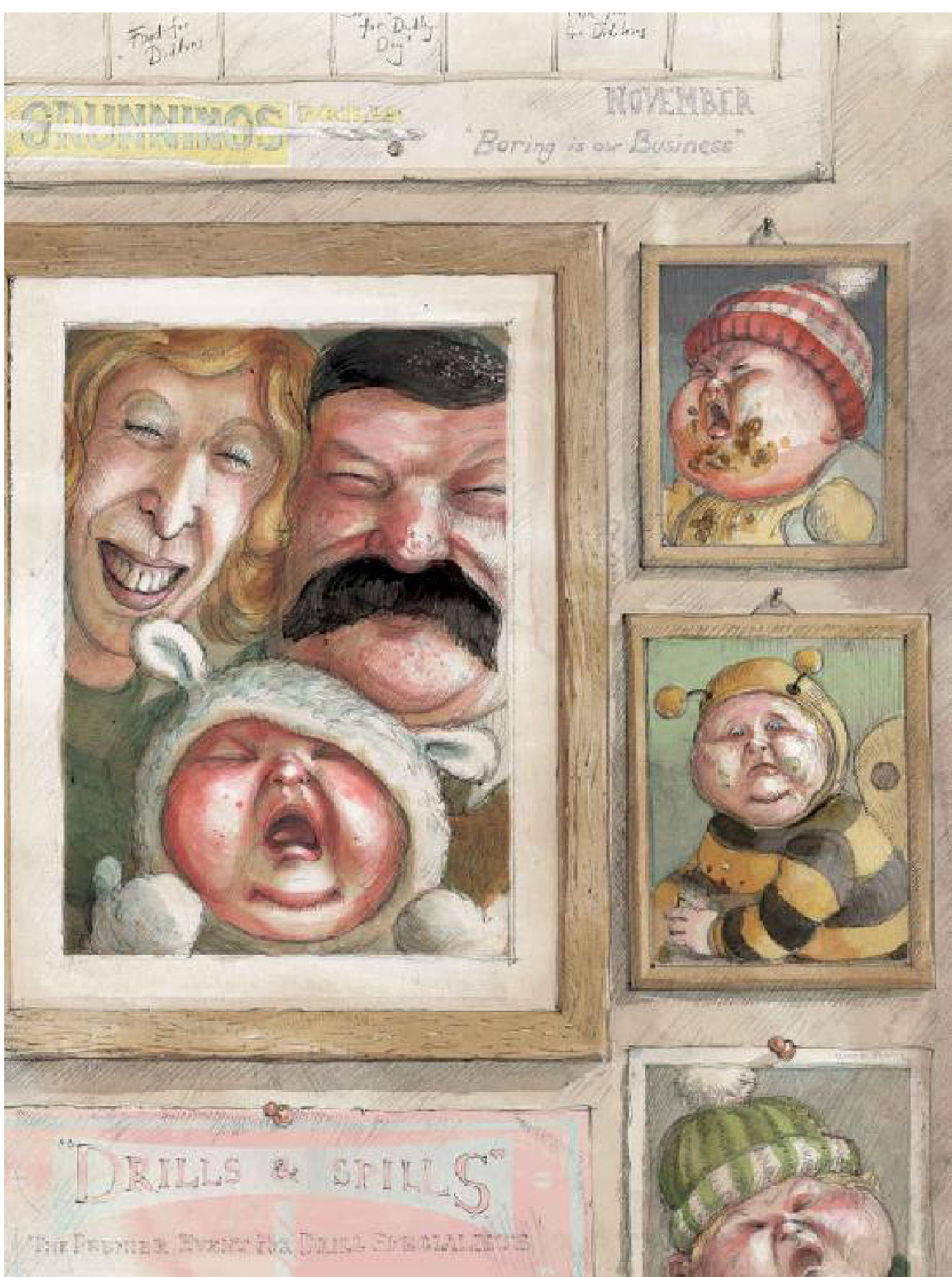}
\caption{The Dursleys' photo wall. From Chapter 1 of "Harry Potter and the Sorcerer's Stone".}
\Description{An example image from the novel domain showing a wall with framed family photos, used to illustrate image-entity annotations in the CMEL dataset.}
\label{Fig.s2}
\end{figure}

The CMEL dataset contains supplementary information including text in Markdown format using MinerU and image information in \nolinkurl{kv_store_image_data}. Figure \ref{Fig.s2} illustrates.

Images numbered by appearance order stored separately in \nolinkurl{kv_store_image_data}. Example:
\begin{lstlisting}[basicstyle=\ttfamily\tiny, breaklines=true, frame=single, rulecolor=\color{black}, language=Python, keywordstyle=\color{blue}, stringstyle=\color{NavyBlue}, commentstyle=\color{green}, showstringspaces=false]
"image_2": {
        "image_id": 2,
        "image_path": "./images/image_2.jpg",
        "caption": [],
        "footnote": [],
        "context": "Mr. Dursley was the director of a firm called Grunnings...",
        "chunk_order_index": 0,
        "chunk_id": "chunk-fb8e5b95ca964e204d9e59caeaf25f09",
        "description": "The image depicts a wall adorned with framed pictures and posters. The central frame contains a family portrait featuring two adults and a baby...",
        "segmentation": true
    }
\end{lstlisting}
Ground truth is stored as JavaScript Object Notation (JSON). Example:
\begin{lstlisting}[basicstyle=\ttfamily\tiny, breaklines=true, frame=single, language=XML, keywordstyle=\color{red}, stringstyle=\color{NavyBlue}, commentstyle=\color{green}, rulecolor=\color{black}, showstringspaces=false]
"image_2": [
        {
            "merged_entity_name": "BABY IN RED HAT",
            "description": "A small framed picture of a baby wearing a red hat with a sad expression...",
            "source_image_entities": [
                "BABY IN RED HAT"
            ],
            "source_text_entities": [
                "DUDLEY"
            ]
        },
       ...
    ]
\end{lstlisting}
The evaluation metrics in this dataset are defined as follows:

\begin{equation}
\text{Micro-Accuracy} = \frac{\sum_{i=1}^{N} \text{Correct}_i}{\sum_{i=1}^{N} \text{Total}_i}
\end{equation}

\begin{equation}
\text{Macro-Accuracy} = \frac{1}{M} \sum_{j=1}^{M} \frac{\text{Correct}_j}{\text{Total}_j}
\end{equation}

where $N$ is the number of entities and $M$ is the number of documents.

\subsection{Construction of the CMEL Dataset}
The CMEL construction pipeline is designed to balance scalability and annotation reliability. We first build text and image knowledge graphs from raw documents, then apply iterative quality control to remove duplicate entities and stabilize cross-modal alignment. This process allows us to preserve domain diversity while keeping annotation consistency across the three data sources.

In particular, the workflow alternates between automatic generation and human-in-the-loop verification. LLMs provide high-recall candidates for merging and alignment, while manual checks focus on error-prone cases, such as ambiguous mentions and near-duplicate entities. This strategy improves both coverage and final label accuracy without requiring exhaustive manual annotation.

\noindent\textbf{Step 0: Document Collection.} The documents for the news and academia domains in the CMEL dataset are sourced from the DocBench dataset\cite{zou2024docbench}. As for the novel domain, we choose several works especially suitable for the CMEL task. Specifically:

\begin{itemize}
\small
\item In the academia domain, the papers come from arXiv, focusing on the top-k most cited papers in the natural language processing field on Google Scholar. 
\item In the news domain, the documents are collected from the front page scans of The New York Times, covering dates from February 22, 2022, to February 22, 2024. 
\item For the novel domain, four illustrated novels with a large number of images were collected and split into 14 documents with approximately the same number of pages to facilitate knowledge graph construction and manual inspection. 
\end{itemize}

\noindent\textbf{Step 1: Indexing.} In this step, we follow the process introduced in Methodology to construct the initial knowledge graphs for the raw documents, including both text-based and image-based knowledge graphs. The specific operations are as follows:

\begin{itemize}
\small
\item \textbf{Text-Based Knowledge Graph Construction:} First, text information is extracted from the raw PDF documents and chunked (fixed token sizes). Each text chunk is converted into a knowledge graph using LLMs, and stored in the file \nolinkurl{kv_store_chunk_knowledge_graph.json}. 
\item \textbf{Image-Based Knowledge Graph Construction:} Image knowledge graph stored in \nolinkurl{kv_store_image_knowledge_graph.json} with entity information in \nolinkurl{kv_store_image_data.json}.
\item \textbf{Data Cleaning and Preparation:} Before storing the data, the working directory is cleaned, retaining only necessary files to ensure the cleanliness of the data storage. 
\end{itemize}

\noindent\textbf{Step 2: Check 1.} In this step, LLM is used to determine whether there are any duplicate entities between different text chunks, assisting with manual inspection and corrections. The specific operations are as follows:

\begin{itemize} 
\small
\item \textbf{Adjacency Entities Extraction:} The \nolinkurl{get_all_neighbors} function is used to extract adjacent entities associated with each text chunk to identify potential duplicate entities. 
\item \textbf{Entity Merge Prompt Generation:} Based on the content and entities of each text chunk, generate specific prompt. Then utilize LLM to determine whether these entities might be duplicates and provide suggestions for merging.
\item \textbf{Manual Inspection:} The results from the LLM are manually reviewed to identify any duplicate entities and to edit \nolinkurl{merged_entities.json}, which serves as guides for the next step. 
\end{itemize}

\noindent\textbf{Step 3: Merging.} After manual inspection, the entity merging phase begins, updating the entities and relations in the knowledge graph based on the confirmed results. The specific operations are as follows:

\begin{itemize}
\small
\item \textbf{Entity Name Standardization:} All entity names are standardized to avoid matching issues caused by case differences. 
\item \textbf{De-duplication and Fusion:} The duplicate entities and relations are removed through the merge results, ensuring each entity appears only once in the graph, while updating the description of each merged entity. 
\item \textbf{Knowledge Graph Update:} The merged entities and relations are stored into the respective knowledge graphs, ensuring that the entities and relationships are unique and standardized. 
\end{itemize}

The prompt for entity merging is as follows:
\begin{tcolorbox}[colframe=gray!180, colback=white, coltitle=white, title=Prompt for Finding Duplicate Entities, sharp corners=south, fontupper=\footnotesize\sffamily, before upper={\setlength{\parindent}{1em}\noindent}]

You are an information processing expert tasked with determining whether multiple entities represent the same object and merging the results. Below are the steps for your task:

1.You will receive a passage of text, a list of entities extracted from the text, and each entity's corresponding type and description.

2.Your task is to determine, based on the entity names, types, descriptions, and their contextual relationships in the text, which entities actually refer to the same object.

3.If you identify two or more entities as referring to the same object, merge them into a unified entity record:

    entity\_name: Use the most common or universal name (if there are aliases, include them in parentheses).
    entity\_type: Ensure category consistency.
    description: Combine the descriptions of all entities into a concise and accurate summary.
    source\_entities: Include all entities that were merged into this entity record.
    
4.The output should contain only merged entities—entities that represent the same object and have been merged. Do not include any entity records for entities that were not merged.

-Input-

Passage:

A passage providing contextual information will be given here.

Entity List:

[{"entity\_name": "Entity1", "entity\_type": "Category1", "description": "Description1"},
  {"entity\_name": "Entity2", "entity\_type": "Category2", "description": "Description2"},
  ...]

-Output-

[{"entity\_name": "Unified Entity Name", "entity\_type": "Unified Category", "description": "Combined Description", "source\_entities": ["Entity1", "Entity2"]},...]

-Considerations for Judgment-

1.Name Similarity: Whether the entity names are identical, commonly used aliases, or spelling variations.

2.Category Consistency: Whether the entity categories are consistent or highly related.

3.Description Relevance: Whether the entity descriptions refer to the same object (e.g., overlapping functions, features, or semantic meaning).

4.Contextual Relationships: Using the provided passage, determine whether the entities refer to the same object in context.

\end{tcolorbox}

\noindent\textbf{Step 4: Generation.} In this step, LLM is used to generate the final alignment results, i.e., the alignment between image entities and text entities. The specific operations are as follows:

\begin{itemize}
\small
\item \textbf{Image and Text Entity Alignment:} The LLM analyzes the entity information in the images and aligns it with the entities in the text chunks. The matching results for each image entity with the corresponding text entity are generated. 
\item \textbf{Generation of Final Results:} The generated alignment results are saved as \nolinkurl{aligned_text_entity.json} files, ensuring that the entity information between the images and text is accurately aligned. 
\end{itemize}

\noindent\textbf{Step 5: Check 2.} After generating the results, potential hallucination errors generated by the LLM (such as incorrect entity alignments) need to be screened and corrected. The specific operations are as follows:

\begin{itemize}
\small
\item \textbf{Error Screening:} Check the alignment results generated by the LLM to identify any errors in the fused entity pairs. Ensure that entities requiring fusion actually exist. 
\item \textbf{Random Check:} A random sample comprising 20\% of the data is manually reviewed to evaluate both the completeness and accuracy of the entity fusion process. Completeness refers to the proportion of entities that required fusion and were successfully merged, while accuracy pertains to the correctness of the merged entities. The results are shown in Table \ref{Tab.s1}. 
\end{itemize}

The prompt for fusion is as follows
\begin{tcolorbox}[colframe=gray!180, colback=white, coltitle=white, title=Prompt for Entity Fusion, sharp corners=south, fontupper=\footnotesize\sffamily, before upper={\setlength{\parindent}{1em}\noindent}]

-Task-

Merge the text entities extracted from images and the entities extracted from nearby text (chunks). The two sets of entities should be merged based on context, avoiding duplication, and ensuring that each merged entity is derived from both image entities and text entities.

-Explanation-

1.Analyze the entities from the image and the entities from the nearby text, identifying which ones share overlapping or complementary context.

2.Merge entities only if there is a clear contextual link between them (e.g., they describe the same object, concept, or entity). Avoid creating a merged entity if it does not involve contributions from both sources.

3.For each pair of entities that are merged, output the unified entity name, category, the integrated description, and the original sources of the entities involved.

4.Discard entities that cannot be meaningfully merged (i.e., if no matching entity exists in the other source).

-Input Format-

Image Entities:

[{"entity\_name": "Entity1", "entity\_type": "Category1", "description": "Description1"},

{"entity\_name": "Entity2", "entity\_type": "Category2", "description": "Description2"},
...]

Original Text:

[Here is a paragraph of text that provides context for the reasoning.]

Nearby Text Entities:

[{"entity\_name": "Entity3", "entity\_type": "Category3", "description": "Description3"},

{"entity\_name": "Entity4", "entity\_type": "Category4", "description": "Description4"},
...]

-Output Format-

[{"entity\_name": "Unified Entity Name", "entity\_type": "Category", "description": "Integrated Description", "source\_image\_entities": ["Entity1"], "source\_text\_entities": ["Entity2"]},
...
]
\end{tcolorbox}

The example provided in the prompt is intended to illustrate the task requirements and assist the LLM in understanding the specific objectives. And the example in the prompt is as follows:

\begin{lstlisting}[basicstyle=\ttfamily\footnotesize,
    breaklines=true,
    language=XML,
    frame=single,
    rulecolor=\color{black},
    stringstyle=\color{NavyBlue}, % 字符串颜色
    showstringspaces=false
    ]
-Example Input-
Image Entities:
[
{"entity_name": "Electric Sedan", "entity_type": "Product", "description": "A high-end electric car focusing on performance and design"}
]

Original Text:
Tesla has a leading position in the global electric car market, with its Model S being a luxury electric vehicle equipped with advanced autonomous driving technology and excellent range.

Nearby Text Entities:
[
{"entity_name": "Tesla", "entity_type": "Company", "description": "A well-known American electric car manufacturer"},
{"entity_name": "Model S", "entity_type": "Product", "description": "A luxury electric vehicle released by Tesla"}
]

-Example Output-
{"merged_entity_name": "Model S", "entity_type": "Product", "description": "Model S is a luxury electric vehicle released by Tesla, equipped with advanced autonomous driving technology and excellent range.", "source_image_entities": ["Electric Sedan"], "source_text_entities": ["Model S"]}
\end{lstlisting}

The CMEL dataset evaluates the alignment step rather than the final fused entity representation. In the full MMGraphRAG pipeline, alignment identifies image--text entity pairs, while fusion uses these aligned pairs to merge and update the multimodal graph. Therefore, CMEL focuses on whether entities that should be fused are correctly aligned.

\begin{table}[htbp]
    \centering
    \renewcommand{\arraystretch}{1.1}
    \setlength{\tabcolsep}{2pt}
    \renewcommand{\arraystretch}{1}
    \scalebox{0.95}{
    \begin{tabular}{ccccc}
        \toprule
         \multirow{2}{*}{\textbf{Performance}} & 
         \multicolumn{3}{c}{\textbf{Type}} &
         \multirow{2}{*}{\textbf{Total.}(196)} \\
             \cmidrule(lr){2-4}
                    & News(26) & Aca.(61) & Nov.(109)\\
        \midrule
        \textbf{Coverage} & 96.7 & 98.0 & 97.1 & 97.7 \\
        \textbf{Accuracy} & 100 & 99.1 & 98.4 & 99.0 \\
        \bottomrule
    \end{tabular}}
    \caption{Manual Inspection Results}
    \label{Tab.s1}
\end{table}

\subsection{Complete Results of the CMEL Dataset}
In the fusion experiment, we selected different models for testing. The embedding-based similarity methods used three models: all-MiniLM-L6-v2\cite{yin2024study} (MLM), bge-m3\cite{chen2024bge} (BGE), and stella-en-1.5B-v5\cite{zhang2024jasper} (Stella). For the LLMs, we chose Llama3.1-70B-Instruct\cite{llama3.1-70b-instruct} (L) and Qwen2.5-72B-Instruct\cite{yang2024qwen2} (Q), while for the MLLMs, we selected Qwen2-VL-72B\cite{wang2024qwen2} (Qvl) and InternVL2.5-38B-MPO\cite{chen2024internvl} (Intvl). 

For the clustering-based approach, the embedding model employed was uniformly Stella-EN-1.5B-V5. After clustering, CMEL requires selecting the appropriate cluster for the target image entity, with two specific methods: k-nearest neighbors (KNN)\cite{guo2003knn} (K) and LLM-based judgment (L). The LLM utilized was Qwen2.5-72B-Instruct.

The displayed experimental results represent the best outcomes from three runs, and the full results are presented in Table \ref{Tab.s2}.

\begin{table}[htbp]
    \centering
    \renewcommand{\arraystretch}{1.1}
    \setlength{\tabcolsep}{5pt}
    \scalebox{0.9}{
    \begin{tabular}{ccccc}
        \toprule
        \multirow{4}{*}{\textbf{Method}} &       
        \multicolumn{3}{c}{\textbf{micro/macro Acc.}} &
         \multirow{2}{*}{\textbf{Overall.}}\\
            \cmidrule(lr){2-4}
                & News & Aca. & Nov.\\        \midrule
        \multicolumn{5}{c}{\textbf{Embedding Model-based Methods}} \\
        \textbf{MLM} & 2.2/1.7 & 15.4/14.9 & 3.9/2.8 & 9.0/6.5 \\
        \textbf{BGE} & 6.5/5.7 & 26.9/26.5 & 9.3/8.4 & 17.0/13.5 \\
        \textbf{Stella} & 10.8/8.4 & 33.1/34.5 & 9.0/7.5 & 20.0/16.8 \\
        \multicolumn{5}{c}{\textbf{LLM-based Methods}} \\
        \textbf{L-Qvl} & 10.8/8.4 & 33.1/34.5 & 9.0/7.5 & 20.0/16.8 \\
        \textbf{L-Intvl} & 10.8/16.7 & 30.2/30.0 & 13.5/13.3 & 20.8/16.8 \\
        \textbf{Q-Qvl} & 31.2/24.1 & 32.2/33.3 & 19.4/23.2 & 26.1/26.8 \\
        \textbf{Q-Intvl} & 33.3/24.1 & 36.8/36.1 & 17.4/20.8 & 27.1/27.0 \\
        \multicolumn{5}{c}{\textbf{Clustering-based Methods}} \\
        \textbf{DB-K} & 48.4/43.1 & 57.0/58.4 & 29.9/31.3 & 43.5/44.3 \\
        \textbf{DB-L} & 53.8/45.9 & 60.8/58.3 & 29.9/34.2 & 45.2/46.1 \\
        \textbf{KM-K} & 48.4/41.5 & 58.2/59.4 & 29.6/29.4 & 43.9/43.4 \\
        \textbf{KM-L} & 50.5/40.6 & 60.7/57.7 & 29.6/30.5 & 45.2/43.0 \\
        \textbf{PR-K} & 50.5/43.0 & 61.0/56.4 & 29.2/33.2 & 44.7/44.2 \\
        \textbf{PR-L} & 51.6/44.4 & 59.7/56.8 & 29.1/35.2 & 44.1/45.5 \\
        \textbf{Lei-K} & 50.5/42.1 & 66.7/64.3 & 30.4/37.2 & 47.7/47.9 \\
        \textbf{Lei-L} & 54.8/44.7 & 60.5/55.5 & 29.4/30.6 & 44.8/43.6 \\
        \textbf{Spe-K} & \underline{57.5/50.9} & \underline{70.1/66.1} & \underline{31.0/39.8} & \underline{49.7/55.1} \\
        \textbf{Spe-L} & \textbf{65.5/56.9} & \textbf{73.3/69.9} & \textbf{31.2/39.4} & \textbf{51.8/59.2} \\
        \bottomrule
    \end{tabular}}
    \caption{Complete Results for CMEL dataset}
    \label{Tab.s2}
\end{table}

The performance differences across embedding models are quite pronounced: in particular, the relatively weak all-MiniLM-L6-v2 model nearly fails to achieve effective cross-modal entity alignment when using embedding-based similarity methods. The same pattern holds for LLM-based approaches: Llama 3.1-70B-Instruct performs significantly worse than Qwen 2.5-72B-Instruct in both the news and novel domains, indicating that CMEL task performance is heavily influenced by model architecture and capability. By contrast, clustering-based methods yield more consistent results on the CMEL task. Overall, assigning categories to image entities via LLMs outperforms KNN-based assignment, although the difference is modest. Moreover, the KNN approach requires one fewer model invocation and runs faster, so either method can be chosen flexibly depending on practical considerations and domain requirements.
\section{Complete Results of Multimodal DocQA Experiments}
\begin{table}[!htbp]
    \centering
    \renewcommand{\arraystretch}{1.1}
    \setlength{\tabcolsep}{2pt}
    \scalebox{0.84}{
    \begin{tabular}{cccccccccc}
        \toprule
         \multirow{2}{*}{\textbf{Model}} &
         \multicolumn{5}{c}{\textbf{Type}} &
         \multicolumn{3}{c}{\textbf{Domain}} &
         \multirow{2}{*}{\textbf{\makecell[c]{Overall \\ Acc.}}}\\
            \cmidrule(lr){2-6}
            \cmidrule(lr){7-9}
                & Aca. & Fin. & Gov. & Laws & News & Text. & Multi. & Una.\\        
        \midrule
        \multicolumn{10}{c}{\textbf{LLM-based Methods}} \\
        \textbf{Llama} & 43.9 & 13.5 & 53.4 & 44.5 & \underline{79.7} & 52.9 & 18.8 & \textbf{81.5} & 44.7 \\
        \textbf{Qwen} & 41.3 & 16.3 & 50.7 & 49.7 & 77.3 & 53.9 & 20.1 & 75.8 & 44.8 \\
        \textbf{Mistral} & 32.3 & 13.2 & 43.9 & 36.1 & 58.1 & 43.0 & 14.6 & 70.2 & 36.0 \\
        \multicolumn{10}{c}{\textbf{MMLLM-based Methods}} \\
        \textbf{Ovis} & 16.2 & 11.1 & 23.6 & 25.7 & 39.0 & 22.8 & 8.8 & 54.8 & 21.3 \\
        \textbf{Qvl} & 17.5 & 14.9 & 25.0 & 34.6 & 48.8 & 34.0 & 8.4 & 40.3 & 25.4 \\
        \textbf{Intvl} & 19.8 & 16.3 & 28.4 & 31.4 & 46.5 & 35.7 & 15.9 & 39.5 & 27.7 \\
        \multicolumn{10}{c}{\textbf{NaiveRAG-based Methods}} \\
        \textbf{Llama} & 43.6 & 38.2 & 66.2 & 64.9 & \textbf{80.2} & 79.9 & 32.1 & 70.2 & 61.0 \\
        \textbf{Qwen} & 43.6 & 34.4 & 62.8 & 65.4 & 75.0 & \underline{81.6} & 30.5 & 67.7 & 59.5 \\
        \textbf{Mistral} & 44.9 & 35.4 & 58.1 & 62.3 & 76.7 & 76.5 & 32.1 & 69.4 & 58.6 \\
        \multicolumn{10}{c}{\textbf{GraphRAG-based Methods}} \\
        \textbf{Llama} & 40.6 & 27.1 & 56.8 & 59.7 & 75.0 & 73.5 & 24.4 & \underline{76.6} & 54.7 \\
        \textbf{Qwen} & 39.6 & 25.7 & 52.5 & 49.7 & 74.5 & 71.7 & 26.0 & 67.5 & 52.3 \\
        \textbf{Mistral} & 37.0 & 28.8 & 59.2 & 61.1 & 75.6 & 67.7 & 26.1 & 76.5 & 49.2 \\
        \multicolumn{10}{c}{\textbf{MMGraphRAG-based Methods (Ours)}} \\
        \textbf{L-Ovis} & 49.7 & 43.6 & 58.5 & 60.0 & 75.3 & 75.1 & 62.5 & 76.3 & 64.1 \\
        \textbf{Q-Ovis} & 50.3 & 46.4 & 59.4 & 56.1 & 76.8 & 76.3 & 63.4 & 74.2 & 65.3 \\
        \textbf{M-Ovis} & 47.9 & 40.6 & 58.6 & 59.3 & 75.2 & 72.6 & 58.6 & 74.1 & 60.9 \\
        \textbf{L-Qvl} & 51.8 & 59.4 & 62.8 & 60.7 & 77.9 & 79.1 & 77.8 & 70.2 & 74.0 \\
        \textbf{Q-Qvl} & 51.8 & 62.9 & \textbf{66.9} & \underline{68.6} & 76.2 & \textbf{82.4} & 81.1 & 67.7 & 75.2 \\
        \textbf{M-Qvl} & 48.4 & 52.8 & 57.9 & 62.7 & 74.5 & 77.0 & 75.4 & 69.6 & 73.9 \\
        \textbf{L-Intvl} & \textbf{60.7} & \underline{64.1} & 62.6 & 64.9 & 76.2 & 80.0 & \underline{86.4} & 75.0 & \textbf{77.5} \\
        \textbf{Q-Intvl} & \underline{60.5} & \textbf{65.8} & \underline{66.5} & \textbf{70.4} & 77.1 & 81.2 & \textbf{88.7} & 71.9 & \underline{76.8} \\
        \textbf{M-Intvl} & 56.4 & 58.1 & 58.0 & 60.2 & 75.2 & 76.6 & 84.9 & 73.3 & 75.7 \\
        \bottomrule
        \end{tabular}}
    \caption{Complete Results of DocBench Dataset. Based on the experimental results from a total of 21 combinations of the six models, designating Llama3.1-70B-Instruct as the evaluation model does not show undue favoritism towards its own generated results, thereby avoiding erroneous evaluation outcomes. }
    \label{Tab.s3}
\end{table}

This experiment selects a variety of models as comparison benchmarks, including LLMs such as Llama3.1-70B-Instruct\cite{llama3.1-70b-instruct} (L), Qwen2.5-72B-Instruct\cite{yang2024qwen2} (Q), and Mistral-Large-Instruct-2411\cite{mistral} (M), as well as MLLMs such as Ovis1.6-Gemma2-27B\cite{lu2024ovis} (Ovis), Qwen2-VL-72B\cite{wang2024qwen2} (Qvl), and InternVL2.5-38B-MPO\cite{chen2024internvl} (Intvl).

Among them, Ovis1.6-Gemma2-27B is deployed using the AutoModelForCausalLM from the Transformers\cite{wolf2020transformers} library, InternVL2.5-38B-MPO is deployed using lmdeploy\cite{chen2024internvl}, and the other models are deployed using vllm\cite{kwon2023efficient}.

The complete results for the DocBench dataset are shown in Table \ref{Tab.s3}, the complete results for the MMLongbench dataset are shown in Table \ref{Tab.s4}, and the results for the MMLongBench dataset by domain are shown in Table \ref{Tab.s5}.

\begin{table}[htbp]
    \centering
    \renewcommand{\arraystretch}{1.1}
    \setlength{\tabcolsep}{2pt}
    \scalebox{0.88}{
    \begin{tabular}{cccccccccc}
        \toprule
         \multirow{2}{*}{\textbf{Model}} &         \multicolumn{3}{c}{\textbf{Locations}} &
         \multicolumn{5}{c}{\textbf{Modalities}} &
         \multirow{2}{*}{\textbf{\makecell[c]{Overall \\ Acc.}}} \\
             \cmidrule(lr){2-4}
             \cmidrule(lr){5-9}
                    & Sin. & Mul. & Una. & Cha. & Tab. & Txt. & Lay. & Fig. & \\
        \midrule
        \multicolumn{10}{c}{\textbf{LLM-based Methods}} \\
        \textbf{Llama} & 23.7 & 20.3 & 51.6 & 16.3 & 12.7 & 32.1 & 21.9 & 17.4 & 28.2 \\
        \textbf{Qwen} & 22.5 & 20.0 & 53.2 & 16.8 & 12.6 & \underline{33.3} & 23.5 & 16.1 & 27.8 \\
        \textbf{Mistral} & 19.1 & 19.2 & 38.6 & 15.6 & 9.4 & 28.9 & 19.2 & 16.2 & 23.1 \\
        \multicolumn{10}{c}{\textbf{MMLLM-based Methods}} \\
        \textbf{Ovis} & 10.6 & 9.5 & 13.0 & 6.9 & 3.1 & 10.0 & 15.4 & 15.4 & 10.7 \\
        \textbf{Qvl} & 10.8 & 9.9 & 8.1 & 7.6 & 5.4 & 10.0 & 8.8 & 12.7 & 10.0 \\
        \textbf{Intvl} & 13.3 & 7.9 & 13.9 & 8.8 & 8.1 & 10.7 & 11.8 & 11.4 & 11.6 \\
        \multicolumn{10}{c}{\textbf{NaiveRAG-based Methods}} \\
        \textbf{Llama} & 24.9 & 19.5 & 56.5 & 16.3 & 15.3 & 31.0 & 22.7 & 18.7 & 29.2 \\
        \textbf{Qwen} & 22.3 & 16.4 & 52.5 & 15.1 & 10.8 & 30.3 & 20.3 & 14.9 & 26.2 \\
        \textbf{Mistral} & 21.6 & 19.2 & 52.9 & 15.4 & 12.5 & 31.2 & 20.6 & 13.7 & 27.3 \\
        \multicolumn{10}{c}{\textbf{GraphRAG-based Methods}} \\
        \textbf{Llama} & 16.3 & 12.3 & \underline{78.5} & 7.6 & 6.7 & 25.1 & 15.0 & 10.6 & 27.2 \\
        \textbf{Qwen} & 18.2 & 13.2 & 77.1 & 14.0 & 11.0 & 26.1 & 12.2 & 8.5 & 28.1 \\
        \textbf{Mistral} & 13.8 & 10.6 & \textbf{86.5} & 9.0 & 5.2 & 21.7 & 13.4 & 7.6 & 27.2 \\
        \multicolumn{10}{c}{\textbf{MMGraphRAG-based Methods (Ours)}} \\
        \textbf{L-Ovis} & 36.9 & 13.0 & 57.4 & 24.4 & 23.7 & 26.8 & 15.9 & 24.1 & 31.0 \\
        \textbf{Q-Ovis} & 37.4 & 15.5 & 54.4 & 24.9 & 23.3 & 28.0 & 26.1 & 27.6 & 31.6 \\
        \textbf{M-Ovis} & 34.7 & 12.0 & 60.0 & 25.6 & 21.2 & 25.7 & 14.8 & 22.1 & 29.5 \\
        \textbf{L-QVL} & 37.6 & 13.8 & 55.2 & 26.4 & 29.2 & 26.4 & 13.8 & 21.2 & 32.6 \\
        \textbf{Q-Qvl} & \underline{38.7} & 20.1 & 51.6 & 26.2 & 30.0 & 29.3 & \textbf{29.1} & \underline{29.2} & 34.8 \\
        \textbf{M-Qvl} & 36.4 & 12.1 & 56.8 & 27.6 & 27.7 & 23.2 & 11.8 & 19.8 & 31.7 \\
        \textbf{L-Intvl} & 38.7 & \underline{21.9} & 59.2 & \textbf{34.7} & \textbf{36.6} & 31.9 & 12.9 & 28.5 & \underline{36.9} \\
        \textbf{Q-Intvl} & \textbf{39.6} & \textbf{26.7} & 55.8 & \underline{34.7} & \underline{36.5} & \textbf{33.8} & \underline{28.7} & \textbf{34.6} & \textbf{38.8} \\
        \textbf{M-Intvl} & 37.7 & 19.5 & 63.2 & 35.0 & 33.5 & 28.0 & 13.6 & 27.6 & 35.7 \\
        \bottomrule
    \end{tabular}}
    \caption{Complete Results of MMLongBench Dataset. The GraphRAG-based method has a similar overall accuracy to the NaiveRAG-based method because it performs better on unanswerable questions but worse on answerable multimodal questions. The MMGraphRAG-based method achieves the best overall accuracy, indicating that it can better answer general questions while maintaining robust performance on unanswerable questions.}
    \label{Tab.s4}
\end{table}

Through comparative experiments with other single-modal and multi-modal models under various methods, we found that Llama3.1-70B-Instruct is capable of maintaining a degree of independence and objectivity during the evaluation process. This suggests that its evaluation mechanism can effectively distinguish the generation results of different models without bias arising from its own model origin. Therefore, it can be concluded that the evaluation conclusions based on Llama3.1-70B-Instruct are relatively reliable and can provide fair and accurate assessment results in multi-model document question answering (QA) experiments.

\begin{table}[H]
    \centering
    \renewcommand{\arraystretch}{1.1}
    \setlength{\tabcolsep}{2pt}
    \scalebox{0.88}{
    \begin{tabular}{ccccccccc}
        \toprule
         \multirow{2}{*}{\textbf{Model}} &  
         \multicolumn{7}{c}{\textbf{Evidence Locations}} &
         \multirow{2}{*}{\textbf{\makecell[c]{Overall \\ Acc.}}} \\ 
             \cmidrule(lr){2-8}
                    & Int. & Tut. & Aca. & Gui. & Bro. & Adm. & Fin. & \\
        \midrule
        \multicolumn{9}{c}{\textbf{LLM-based Methods}} \\
        \textbf{Llama} & 33.5 & 32.1 & 27.7 & 24.7 & 22.0 & 31.1 & 18.8 & 28.2 \\
        \textbf{Qwen} & 31.5 & 31.8 & 25.4 & 30.3 & 26.5 & \textbf{36.6} & 9.6 & 27.8 \\
        \textbf{Mistral} & 28.2 & 25.7 & 19.3 & 22.0 & 22.9 & 32.8 & 8.4 & 23.1 \\
        \multicolumn{9}{c}{\textbf{MMLLM-based Methods}} \\
        \textbf{Ovis} & 7.4 & 20.4 & 10.9 & 7.3 & 16.5 & 14.3 & 4.3 & 10.7 \\
        \textbf{Qvl} & 7.8 & 20.1 & 6.8 & 10.6 & 9.8 & 11.6 & 7.2 & 10.0 \\
        \textbf{Intvl} & 11.3 & 19.1 & 10.4 & 8.9 & 14.4 & 13.3 & 6.0 & 11.6 \\
        \multicolumn{9}{c}{\textbf{NaiveRAG-based Methods}} \\
        \textbf{Llama} & 34.0 & 31.0 & 30.0 & 29.4 & 23.0 & 29.6 & 18.4 & 29.2 \\
        \textbf{Qwen} & 30.0 & 27.6 & 25.9 & 31.1 & 17.0 & 32.1 & 12.8 & 26.2 \\
        \textbf{Mistral} & 32.6 & 24.7 & 24.5 & 32.5 & 18.7 & 34.0 & 17.6 & 27.3 \\
        \multicolumn{9}{c}{\textbf{GraphRAG-based Methods}} \\
        \textbf{Llama} & 30.8 & 27.0 & 25.0 & 29.7 & 24.0 & 34.4 & 16.7 & 27.2 \\
        \textbf{Qwen} & 34.6 & 21.8 & 24.8 & 29.2 & 27.0 & \underline{36.2} & 18.9 & 28.1 \\
        \textbf{Mistral} & 34.4 & 22.3 & 24.9 & 28.1 & 26.0 & 31.6 & 15.5 & 27.2 \\
        \multicolumn{9}{c}{\textbf{MMGraphRAG-based Methods (Ours)}} \\
        \textbf{L-Ovis} & 35.7 & 32.9 & 30.2 & \underline{41.7} & 27.1 & 24.6 & 26.7 & 31.0 \\
        \textbf{Q-Ovis} & 36.4 & \textbf{38.9} & 29.4 & 39.3 & \underline{32.2} & 30.2 & 30.1 & 31.6 \\
        \textbf{M-Ovis} & 36.5 & 34.2 & 29.4 & 38.3 & 31.5 & 22.5 & 25.5 & 29.5 \\
        \textbf{L-Qvl} & 36.5 & 32.9 & 27.1 & 41.3 & 26.3 & 28.4 & 34.0 & 32.6 \\
        \textbf{Q-Qvl} & 37.2 & \underline{38.8} & 26.3 & 39.5 & 31.6 & 35.8 & \underline{38.0} & 34.8 \\
        \textbf{M-Qvl} & 37.0 & 35.2 & 26.5 & 38.8 & 30.9 & 26.4 & 32.2 & 31.7 \\
        \textbf{L-Intvl} & 42.4 & 34.0 & \textbf{36.8} & \textbf{42.8} & 28.9 & 26.0 & 36.2 & \underline{36.9} \\
        \textbf{Q-Intvl} & \textbf{43.3} & 36.5 & 35.2 & 40.0 & \textbf{32.7} & 33.5 & \textbf{41.0} & \textbf{38.8} \\
        \textbf{M-Intvl} & \underline{42.9} & 35.0 & \underline{35.6} & 38.9 & 31.1 & 23.2 & 35.2 & 35.7 \\
        \bottomrule
    \end{tabular}}
    \caption{MMLongBench Dataset Domain Results. The MMGraphRAG method achieves the best performance across all domains except for the Administration/Industry file category. Notably, it demonstrates the most significant improvements in the Guidebook and Financial report categories, which are characterized by a high volume of charts ,tables and figures. These enhancements are far more pronounced than those of other methods.}
    \label{Tab.s5}
\end{table}

The MMLongBench dataset encompasses a diverse range of document domains. These domains include Research Reports/Introductions (Int.), which typically feature academic or industry-oriented analyses and background information; Tutorials/Workshops (Tut.), focusing on instructional content for skill development or knowledge dissemination; Academic Papers (Aca.), containing scholarly research and findings; Guidebooks (Gui.), offering practical information and advice for specific topics or activities; Brochures (Bro.), designed for promotional or informational purposes in a concise format; Administration/Industry Files (Adm.), covering official documents or industry-specific reports; and Financial Reports (Fin.), presenting financial data and analyses. 
\section{Generation Module}
Given that the current technology for MLLMs is still not fully mature, the Generation module employs a hybrid-generation approach that combines results from an LLM (text-only model) and an MLLM (capable of processing text, images, and other multi-modal information). This strategy aims to enhance the accuracy and stability of the final output.

This hybrid strategy not only makes full use of the maturity and advantages of LLMs in specific tasks, ensuring high quality semantic understanding and language expression, but also enriches the generated content by integrating multi-modal information. This approach effectively mitigates the limitations of current MLLMs while ensuring an overall improvement in the quality of the system's generation.

The Generation module achieves hybrid-generation through five steps: First, it extracts relevant information from the output of retrieval module and generates an initial textual response using LLM. Next, the MLLM is invoked to combine image and text information to generate a multimodal response, which is then merged with other multimodal reponses through an LLM. Finally, the generated textual response and multimodal response are integrated to output a comprehensive and consistent final result.

A specific example of an execution process is as follows, with the user's query and the generated answer provided:

\noindent \textbf{question:} "What is the highest F1 score achieved on the Chinese OntoNotes4.0 dataset, according to Table 10?"

\textbf{answer:} "The highest F1 score on the Chinese OntoNotes4.0 dataset is 84.67."

\textbf{type:} "multi-modal-t"

\textbf{evidence:} "The question requires identifying the maximum value in a single column, which is straightforward."

\noindent \textbf{LLM Response:}

\begin{quote}
\textit{\textcolor[HTML]{5D95DA}{\small The information cannot be obtained from the provided data tables.}}
\end{quote}

\noindent \textbf{MLLM Responses:}

{\centering response 1 \par}

\begin{quote}
\textit{\textcolor{gray}{\small To answer the question about the highest F1 score achieved on the Chinese OntoNotes 4.0 dataset, we will refer to the table labeled "Table 10" in the image provided. This table displays the effects of varying hyperparameters, specifically the alpha values, on the performance metrics of two datasets: Chinese OntoNotes 4.0 and English QuoRef. The table columns include alpha values ranging from 0.1 to 0.9 in increments of 0.1. For each alpha value, there are corresponding F1 scores for both Chinese OntoNotes 4.0 and English QuoRef datasets. Upon examining the "Chinese OntoNotes 4.0" column in the table, we notice the following F1 scores for each alpha value:- alpha = 0.1: F1 score = 80.13 - alpha = 0.2: F1 score = 81.17 - alpha = 0.3: F1 score = 84.22 - alpha = 0.4: F1 score = 84.52 - alpha = 0.5: F1 score = 84.47 - alpha = 0.6: F1 score = 84.67 (bolded, indicating the highest score) - alpha = 0.7: F1 score = 81.81 - alpha = 0.8: F1 score = 80.97 - alpha = 0.9: F1 score = 80.21 } \textcolor[HTML]{B02418}{\small From this data, it is clear that the highest F1 score for the Chinese OntoNotes 4.0 dataset is **84.67**, which is achieved at an alpha value of 0.6. This score is highlighted in bold in the table, further confirming its significance as the peak performance metric for this specific dataset within the given range of alpha values.}}
\end{quote}

{\centering response 2 \par}

\begin{quote}
\textit{\textcolor[HTML]{B02418}{\small According to the table in image 12, the highest F1 score achieved on the Chinese OntoNotes 4.0 dataset is 84.67, which occurs at an alpha value of 0.6. This table illustrates the effects of hyperparameters on the performance metrics of the Chinese OntoNotes 4.0 and English QuoRef datasets, showing that the F1 score for Chinese OntoNotes 4.0 peaks at 84.67 when (alpha) is set to 0.6.}}
\end{quote}
 
{\centering response 3 \par}

\begin{quote}
\textit{\textcolor[HTML]{B02418}{\small I'm sorry, but I cannot provide an analysis of the table in the image as it was not provided.}}
\end{quote}

\noindent \textbf{Merged MLLM Response:}

\begin{quote}
\textit{\textcolor[HTML]{E27A33}{\small The highest F1 score achieved on the Chinese OntoNotes 4.0 dataset is 84.67, which occurs at an alpha value of 0.6. This information is confirmed by the data in Table 10, which displays the effects of varying hyperparameters, specifically the alpha values, on the performance metrics of two datasets: Chinese OntoNotes 4.0 and English QuoRef.}}
\end{quote}

\noindent \textbf{Final Result:}

\begin{quote}
\textit{\textcolor[HTML]{2C4D76}{\small The highest F1 score achieved on the Chinese OntoNotes 4.0 dataset, according to analysis of the information provided, is 84.67. This optimal F1 score occurs when the alpha value is set to 0.6.}}
\end{quote}

The prompt for merging multimodal responses is as follows:
\begin{tcolorbox}[colframe=gray!180, colback=white, coltitle=white, title=Prompt for Merging MLLM responses, sharp corners=south, fontupper=\small\sffamily, before upper={\setlength{\parindent}{1em}\noindent}]

The following is a list of responses generated by a multi-modal model based on the same user Query but different images. Please perform the following tasks:

-Analyze the Responses: Identify any contradictions, repetitions, or inconsistencies among the responses.

-Reasonably Determine: Decide which response best aligns with the user Query based on the provided information, ensuring that the determination is based on the relevance and accuracy of the information in the response rather than a majority consensus, as the correct answer may only pertain to a specific image and may not align with the majority.

-Provide a Unified Answer: Deliver a single, unified response that eliminates contradictions, resolves ambiguities, and accurately addresses the user Query.

Additionally, retain any highly relevant information from the responses that supports or complements the unified answer.
\end{tcolorbox}
\section{Implementation of the Fusion Module}

The pseudocode for the entire fusion process is as above, which provides a clearer understanding of the input and output of each step. The final step utilizes the fused MMKG to construct an entity vector database (vdb), which facilitates the retrieval stage. 

$imgdata$ (image data) and $chunks$ (text chunks) are both results of preprocessing, stored in their respective JSON files. $imgdata$ is used to store various information related to images, while $chunks$ store the text chunks of the entire document. $tkg$ (text-based knowledge graph) is the result of the text modality processing module txt2graph. It is stored in JSON files for each chunk and also as a complete GraphML file. Here, we do not make a specific distinction between them. $ikg$ represents the image-based knowledge graph. First, $ikg$ is obtained from the $image$ and $imgdata$. Then, the formal first step is to align the entities in $tkg$ and $ikg$, and save the alignment results as $list_a$. Entities to be fused are filtered out from $ikg$, and the remaining entities are enhanced using $chunks$ (context) and relevant entities from $tkg$, resulting in the enhanced image-based knowledge graph $ikg_e$. 

Next, we perform a detailed search in $tkg$ to align the global entity of the image. This involves extracting relevant text segments from chunks that are semantically related to the whole image. Once the entity is identified, matching is performed in $tkg$. If aligned entity is found, only the relations of $ikg_e$ will be updated to obtain $ikg_u$, which naturally achieves alignment during fusion. If no aligned entity is found in $tkg$, $ikg_e$ will be updated by supplementing a new entity obtained from chunks to form $ikg_u$. Finally, $tkg$ and $ikg_u$ are fused based on the results of $list_a$ to obtain the final $mmkg$ (multimodal knowledge graph).

\begin{algorithm}
\caption{Fusion}
\begin{algorithmic}[1]
    \Function{Fusion}{$\text{images}$}
        \State \textbf{Initialize} $imgdata \gets \text{image\_data.json}$
        \State \textbf{Initialize} $tkg \gets \text{chunk\_knowledge\_graph.json}$
        \State \textbf{Initialize} $chunks \gets \text{text\_chunks.json}$
        \For{each $image \in \text{images}$}
            \If{$mmkg$ exists}
                \State \textbf{continue} to next iteration
            \EndIf
            \State $ikg \gets \Call{Find}{imgdata, image}$
            \State $list^a \gets \Call{align}{tkg, ikg}$
            \State $ikg_e \gets \Call{Enhance}{list^a, ikg, tkg, chunks}$
            \State $ikg_u \gets \Call{Update}{ikg_e, tkg, image, chunks}$
            \State $mmkg \gets \Call{Merge}{ikg_u, tkg, list^a}$
        \EndFor
        \State $vdb \gets mmkg$
        \State \Return $mmkg$, $vdb$
    \EndFunction
\end{algorithmic}
\end{algorithm}

The second step, entity enhancement, is achieved using the reasoning capabilities of LLMs. Based on the context information, entities from $tkg$ are used to supplement visual entities that do not have aligned counterparts. The prompt to enhance entities is as follows:

\begin{tcolorbox}[colframe=gray!180, colback=white, coltitle=white, title=Prompt for Enhancing Image Entities, sharp corners=south, fontupper=\footnotesize\sffamily, before upper={\setlength{\parindent}{1em}\noindent}]
The goal is to enrich and expand the knowledge of the image entities listed in the img\_entity\_list based on the provided chunk\_text. 

The entity\_type should remain unchanged, but you may modify the entity\_name and description fields to provide more context and details based on the information in the chunk\_text.

For each entry in the img\_entity\_list, the following actions should be performed:

1. Modify and enhance the entity\_name if necessary.

2. Expand the description by integrating relevant details and insights from the chunk\_text.

3. Include an original\_name field to capture the original entity name before enhancement.

Ensure the final output is in valid JSON format, only including the list of enhanced entities without any additional text.
\end{tcolorbox}

After generating candidate entities, LLM is used to align visual entities, with the specific prompt as follows:
\begin{tcolorbox}[colframe=gray!180, colback=white, coltitle=white, title=Prompt for Generating Image Feature Block Description, sharp corners=south, fontupper=\footnotesize\sffamily, before upper={\setlength{\parindent}{1em}\noindent}]
You are an expert system designed to identify matching entities based on semantic similarity and context. Given the following inputs:

img\_entity: The name of the image entity to be evaluated.

img\_entity\_description: A description of the image entity.

chunk\_text: Text surrounding the image entity providing additional context.

possible\_image\_matched\_entities: A list of possible matching entities. Each entity is represented as a dictionary with the following fields:

  entity\_name: The name of the possible entity.
  
  entity\_type: The type/category of the entity.
  
  description: A detailed description of the entity.
  
  additional\_info: Additional relevant information about why choose this entity (such as similarity, reason generated by LLM, etc.).
  
-Task-

Using the information provided, determine whether the img\_entity matches any of the entities in possible\_image\_matched\_entities. Consider the following criteria:

1.Semantic Matching: Evaluate the semantic alignment between the img\_entity and the possible matching entities, based on their names, descriptions, and types. Even without a similarity score, assess how well the img\_entity matches the attributes of each possible entity.

2.Contextual Relevance: Use the chunk\_text and img\_entity\_description to assess the contextual alignment between the img\_entity and the possible entity.

-Output-

If a match is found, only return the entity\_name of the best-matching entity.

If no match meets the criteria (e.g., low similarity or poor contextual fit), only output "no match".

Do not include any explanations, reasons, or additional information in the output.
\end{tcolorbox}
\section{Implementation of the Img2Graph Module}
\noindent \textbf{Image Segmentation.} The initial step involves subjecting the input image to segmentation, which entails parsing the image into regions that possess distinct semantic meanings (referred to as image feature blocks). This process is achieved through the image segmentation function in the YOLOv8\cite{jocher2023yolo} model. The segmentation outcomes provide a finer-grained representation of image information, thereby establishing the foundation for subsequent entity extraction and relation modeling. The granularity of segmentation significantly influences the precision of edge delineation within knowledge graphs.

\noindent \textbf{Image Feature Block Description.} Subsequently, the MLLM is employed to generate text descriptions for each image feature block derived from the segmentation results. The generation process of the model is guided by carefully designed prompts, which enable it to produce accurate and semantically rich information. These text descriptions not only construct independent entities for the image modality but also provide a bridge for subsequent alignment with the text modality. Specific prompts are as follows: 

\begin{tcolorbox}[colframe=gray!180, colback=white, coltitle=white, title=Prompt for Generating Image Feature Block Description, sharp corners=south, fontupper=\small\sffamily, before upper={\setlength{\parindent}{1em}\noindent}]
You are a multi-modal model capable of processing image feature blocks and generating detailed descriptions.

Your task is to first determine the category of the given image feature block (object, organism, or person) and then extract the entity's features from it, providing a detailed description.

Note that the entity in the image feature block may not be complete, such as a half-body photo of a person or a partial image of an object.

- If the entity is an object, describe the object's features, including its name, color, shape, size, material, possible function, and other significant characteristics.
  
- If the entity is an organism, describe the features of this organism (animal or plant), including species, name, age, color, shape, size, posture, or structural characteristics.
  
- If the entity is a person, describe the person's features, including gender, skin color, hairstyle, clothing, facial expression, age, and posture.
  
All image feature blocks have a black background, so focus solely on the entity's characteristics, and do not mention "the background is black" in the output.
\end{tcolorbox}

The example in the prompt is derived from the processing results illustrated in Methodology, and the example is as follows: 
\begin{lstlisting}[basicstyle=\ttfamily\footnotesize,
    breaklines=true,
    language=XML,
    frame=single,
    rulecolor=\color{black},
    stringstyle=\color{NavyBlue}, % 字符串颜色
    showstringspaces=false,
    ]
The category of this image feature block is 'person'. The entity features are as follows:
Person Features:
- Gender: Female
- Hairstyle: Long hair, light brown, naturally falling with some hair pinned with a clip
- Eyes: Blue, large and expressive
- Expression: Smiling, appears friendly and joyful
- Age: Appears to be a young woman
- Clothing: Wearing a white shirt with the sleeves rolled up, revealing the wrists; paired with blue overalls, with dark blue straps; light blue sneakers on her feet
- Accessories: Orange shoulder bag on her right shoulder; brown belt tied around the waist
- Holding: Holding a vintage-style camera with both hands, the camera is black and silver, with a large lens, appearing professional
Overall, the character gives off a youthful, lively vibe with a touch of artistic flair.
\end{lstlisting}

\noindent \textbf{Entity and Relation Extraction from the Image.} This step employs an MLLM, guided by prompts, to identify explicit relations (e.g., "girl — girl holding a camera — camera") and implicit relations (e.g., "boy — 
the boy and girl appear to be close, possibly friends or a couple — girl"). The extracted entities and relations provide structured information for the multimodal extension of the knowledge graph. 

Compared to traditional scene graph generation methods, MLLM-based approaches excel at extracting entities and inferring both explicit and implicit relations by leveraging the semantic segmentation and reasoning abilities of MLLMs, resulting in high-precision and fine-grained scene graphs.

The prompt for extracting entity and relation from image is as follows:
\begin{tcolorbox}[colframe=gray!180, colback=white, coltitle=white, title=Prompt for Visual Entity and Relation Extraction, sharp corners=south, fontupper=\footnotesize\sffamily, before upper={\setlength{\parindent}{1em}\noindent}]

Given a raw image, extract the entities from the image and generate detailed descriptions of these entities, while also identifying the relationships between the entities and generating descriptions of these relationships. Finally, output the result in a standardized JSON format. Note that the output should be in English.

-Steps-

1. Extract all entities from the image. 
  
  For each identified entity, extract the following information:
  
  - Entity Name: The name of the entity
    
  - Entity Type: Can be one of the following types: [\{entity\_types\}]
    
  - Entity Description: A comprehensive description of the entity's attributes and actions
    
  - Format each entity as ("entity"\{tuple\_delimiter\} Entity\_Name \{tuple\_delimiter\} Entity\_Type \{tuple\_delimiter\} Entity\_Description
    
2. From the entities identified in Step 1, identify all pairs of (Source Entity, Target Entity) where the entities are clearly related. 
  
  For each related pair of entities, extract the following information:
  
  - Source Entity: The name of the source entity, as identified in Step 1
    
  - Target Entity: The name of the target entity, as identified in Step 1
    
  - Relationship Description: Explain why the source entity and target entity are related
    
  - Relationship Strength: A numerical score indicating the strength of the relationship between the source and target entities
    
    Format each relationship as ("relationship" \{tuple\_delimiter\} Source\_Entity \{tuple\_delimiter\} Target\_Entity \{tuple\_delimiter\} Relationship\_Description \{tuple\_delimiter\} Relationship\_Strength)
    
3. Return the output as a list including all entities and relationships identified in Steps 1 and 2. Use \{record\_delimiter\} as the list separator.
  
4. Upon completion, output \{completion\_delimiter\}
\end{tcolorbox}

The examples contained within the prompt are excessively lengthy. For illustrative purposes, only a small excerpt is presented here to demonstrate the format, as follows:
\begin{lstlisting}[basicstyle=\ttfamily\footnotesize,
    breaklines=true,
    language=XML,
    frame=single,
    rulecolor=\color{black},
    stringstyle=\color{NavyBlue}, % 字符串颜色
    showstringspaces=false
    ]
("entity"{tuple_delimiter}"Girl"{tuple_delimiter}"person"{tuple_delimiter}"Wearing glasses, dressed in black, holding white and blue objects, smiling at the camera."){record_delimiter} 
("entity"{tuple_delimiter}"Headphones"{tuple_delimiter}"object"{tuple_delimiter}"White headphones on the girl's ears."){record_delimiter} 
...
("relationship"{tuple_delimiter}"Girl"{tuple_delimiter}"Headphones"{tuple_delimiter}"The girl is wearing headphones."{tuple_delimiter}8){record_delimiter} 
("relationship"{tuple_delimiter}"Girl"{tuple_delimiter}"Phone"{tuple_delimiter}"The girl is holding a phone in her hand."{tuple_delimiter}8){record_delimiter} 
...
\end{lstlisting}

\noindent \textbf{Alignment of Image Feature Blocks with Entities.} Based on the extracted visual entities, the feature blocks generated by segmentation are aligned with their corresponding textual entities. This step is accomplished through the recognition and reasoning capabilities of the MLLM. For example, based on the semantic content of the textual entity, "Feature Block 2" is identified as the image of a "boy," and a relation is established in the knowledge graph. This alignment not only connects feature blocks to entities but also strengthens the association between modalities.

The prompt for aligning visual entities is as follows:
\begin{tcolorbox}[colframe=gray!180, colback=white, coltitle=white, title=Prompt for Visual Entity Alignment, sharp corners=south, fontupper=\small\sffamily, before upper={\setlength{\parindent}{1em}\noindent}]
-Objective-

Given an image feature block and its name placeholder, along with entity-description pairs extracted from the original image, determine which entity the image feature block corresponds to and output the relationship with the entity. The output should be in English.

-Steps-

1. Based on the provided entity-description pairs, determine the entity corresponding to the image feature block and output the following information:
  
- Entity Name: The name of the entity corresponding to the image feature block
  
2. Output the relationship between the image feature block and the corresponding entity, and extract the following information:
  
- Image Feature Block Name: The name of the input image feature block
  
- Relationship Description: Describe the relationship between the entity and the image feature block, with the format "The image feature block Image Feature Block Name is a picture of Entity Name."
  
- Relationship Strength: A numerical score representing the strength of the relationship between the image feature block and the corresponding entity
  
  Be sure to include the \{record\_delimiter\} to signify the end of the relationship.
\end{tcolorbox}

The examples saved in the prompt are as follows:
\begin{lstlisting}[basicstyle=\ttfamily\footnotesize,
    breaklines=true,
    language=XML,
    frame=single,
    rulecolor=\color{black},
    stringstyle=\color{NavyBlue}, % 字符串颜色
    showstringspaces=false
    ]
Example 1:
The image feature block is as shown above, and its name is "image_0_apple-0.jpg."
Entity-Description: 
"Apple" - "A green apple, smooth surface, with a small stem."
"Book" - "Three stacked books, red cover, yellow inner pages."
Output:
("relationship"{tuple_delimiter}"Apple"{tuple_delimiter}"image_0_apple-0.jpg"{tuple_delimiter}"The image feature block image_0_apple-0.jpg is a picture of an apple."{tuple_delimiter}7){record_delimiter}
\end{lstlisting}

\noindent \textbf{Global Entity Construction.} Finally, a global entity is constructed for the entire image, serving as a global node in the knowledge graph. This node not only provides supplementary descriptions of the image's global information (e.g., "meet on the bridge") but also enhances the completeness of the knowledge graph through its connections to local entities. Through this step, the knowledge graph can provide multi-level information from global to local, further enhancing retrieval capabilities.
\section{Impact of Model Scale on System Performance}

To characterize how model scale affects MMGraphRAG, we compare small-scale (7B) and large-scale (72B) models across two stages: graph construction and answer generation. The goal is to provide an objective analysis of efficiency and accuracy under different scale allocations.

\subsection{Experimental Setup.}
We evaluate four model-scale configurations: 7B-7B, 7B-72B, 72B-72B, and 72B-7B, where the first model is used for graph construction and the second for answer generation. We use Qwen2.5-7B/72B-Instruct for text processing and Qwen2-VL-7B/72B for multimodal parsing. All experiments are conducted on DocBench.

All models are deployed on the same Ubuntu server with vLLM and two NVIDIA A6000 GPUs, without quantization, to ensure comparable runtime and accuracy across configurations; all other settings remain unchanged.

We use open-weight model families to make the pipeline reproducible and deployable in privacy-sensitive settings. Many target scenarios involve internal reports, enterprise document stores, or regulated documents that cannot be sent to third-party application programming interface (API) services. Open models also allow fixed versions, local batching, and stable re-execution, so changes in performance can be attributed to graph construction and model scale rather than to opaque API updates.

\subsection{Graph Construction Runtime.}

\begin{table}[tbp]
    \centering
    \renewcommand{\arraystretch}{1.2}
    \setlength{\tabcolsep}{2pt}
    \resizebox{\columnwidth}{!}{%
    \begin{tabular}{lcccccc}
        \toprule
        \multirow{2}{*}{\textbf{Config}} &
        \multicolumn{5}{c}{\textbf{Graph Construction Time (minutes)}} \\
        \cmidrule(lr){2-6}
        & \textbf{Aca.} & \textbf{Fin.} & \textbf{Gov.} & \textbf{Law.} & \textbf{New.} \\
        \midrule
        \textbf{7B} & 28 (20--47) & 128 (25--280) & 74 (14--170) & 68 (22--134) & 3 (2--4) \\
        \textbf{72B} & 19 (14--31) & 86 (17--188) & 50 (9--114) & 46 (15--90) & 2 (1--2) \\
        \bottomrule
    \end{tabular}%
    }
    \caption{Graph construction time by model scale and document category. Values are reported as average time with min-max range (minutes).}
    \label{tab:graph_construction_time}
\end{table}

As shown in Table~\ref{tab:graph_construction_time}, the 72B model is consistently faster than the 7B model in graph construction, with about $1.5\times$ speedup on average (e.g., 28 vs. 19 minutes for academia).

Runtime varies across domains due to document size and structure. Finance documents incur the highest cost, while News documents (single-page) complete within a few minutes. Overall, runtime remains bounded and scales sub-linearly due to chunked and partially parallel processing.

\subsection{Accuracy Results.}

\begin{table}[tbp]
    \centering
    \setlength{\tabcolsep}{2pt}
    \scalebox{0.95}{
    \begin{tabular}{lcccccccccc}
        \toprule
        \multirow{2}{*}{\textbf{Config}} &
        \multicolumn{5}{c}{\textbf{Domains (\%)}} &
        \multicolumn{3}{c}{\textbf{Modalities (\%)}} &
        \multirow{2}{*}{\textbf{\makecell[c]{Overall \\ Acc.}}}\\
        \cmidrule(lr){2-6}
        \cmidrule(lr){7-9}
        & Aca. & Fin. & Gov. & Law. & New. & Text & Multi & Una. & \\
        \midrule
        \textbf{7B-7B} & 37.6 & 44.6 & 48.6 & 47.6 & 58.7 & 66.5 & 54.7 & 57.9 & 55.3 \\
        \textbf{7B-72B} & 40.4 & 48.4 & 52.9 & 52.3 & 62.5 & 70.0 & 60.8 & 60.9 & 60.2 \\
        \textbf{72B-72B} & 51.8 & 62.9 & 66.9 & 68.8 & 76.2 & 82.4 & 81.1 & 67.7 & 75.2 \\
        \textbf{72B-7B} & 48.2 & 57.9 & 61.5 & 62.6 & 71.6 & 78.3 & 73.0 & 64.3 & 69.2 \\
        \bottomrule
    \end{tabular}}
    \caption{Answer accuracy under different model-scale configurations on DocBench.}
    \label{tab:accuracy_model_scales}
\end{table}

Table~\ref{tab:accuracy_model_scales} reports answer accuracy for all scale combinations. The 72B-72B configuration achieves the highest overall accuracy (75.2\%). Replacing the construction model with 7B (7B-72B) reduces overall accuracy to 60.2\%, while replacing only the generation model with 7B (72B-7B) yields 69.2\%.

\paragraph{Analysis.}
These results indicate that model scale in graph construction has a larger impact than model scale in answer generation. A stronger construction model produces higher-quality MMKGs, which improves downstream performance even when generation is handled by a smaller model. The modular design of MMGraphRAG therefore enables flexible scale allocation based on target accuracy and computational budget.
\section{Reasoning Path Interpretability}

To further reveal the mechanisms underlying MMGraphRAG's ability to suppress hallucinations, this section provides a white-box analysis of the graph retrieval and multi-hop reasoning process from the perspectives of interpretability and evidence provenance. We illustrate this using a complex comparative multimodal question from document QA: \textit{"Which model variant has the highest improvement in F1 score for the QuoRef dataset when compared to the base XLNet model?"} We follow a normalized reasoning pipeline: global graph construction $\rightarrow$ subgraph retrieval $\rightarrow$ evidence alignment $\rightarrow$ conclusion generation.

For each query, the system records the retrieved entities, relations, supporting text units, and attached image evidence in \nolinkurl{context.csv}. This file provides the provenance trail used to inspect how the final answer is grounded in the retrieved local subgraph.

\subsection{Structured Provenance Based on Global Graph.}
\begin{figure}[tp]
    \centering
    \includegraphics[width=\linewidth]{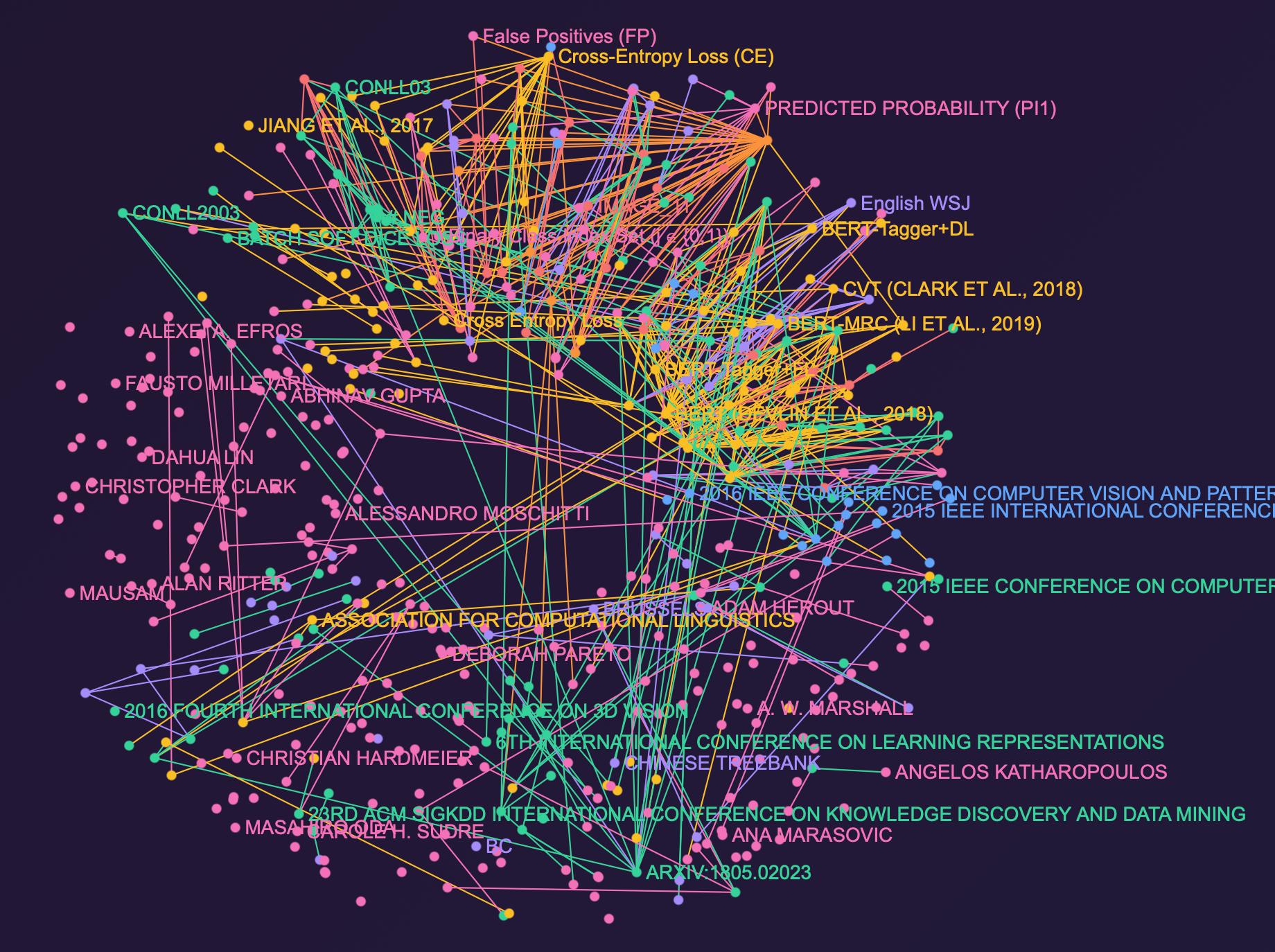}
    \caption{Complete MMKG global topology (macroscopic evidence space before reasoning)}
    \Description{Global topology view of the complete MMKG used before reasoning, showing the full evidence space with connected entities and relations.}
    \label{fig:full-mmkg-q3}
\end{figure}

System interpretability is founded on a complete underlying topological structure. For the document in question, after end-to-end multimodal graph construction, the system generates a global MMKG comprising 527 nodes, 646 edges, and 218 connected components (see Figure~\ref{fig:full-mmkg-q3}). Text entities (415) and visual entities (23) are seamlessly integrated. The increased graph density and expanded node scale indicate that discrete image-text evidence has been successfully woven into a unified logical network, providing a solid structured foundation for subsequent evidence tracking and multi-hop traversal.

\subsection{Query-Driven Local Subgraph Extraction.}

\begin{figure}[tp]
    \centering
    \includegraphics[width=\linewidth]{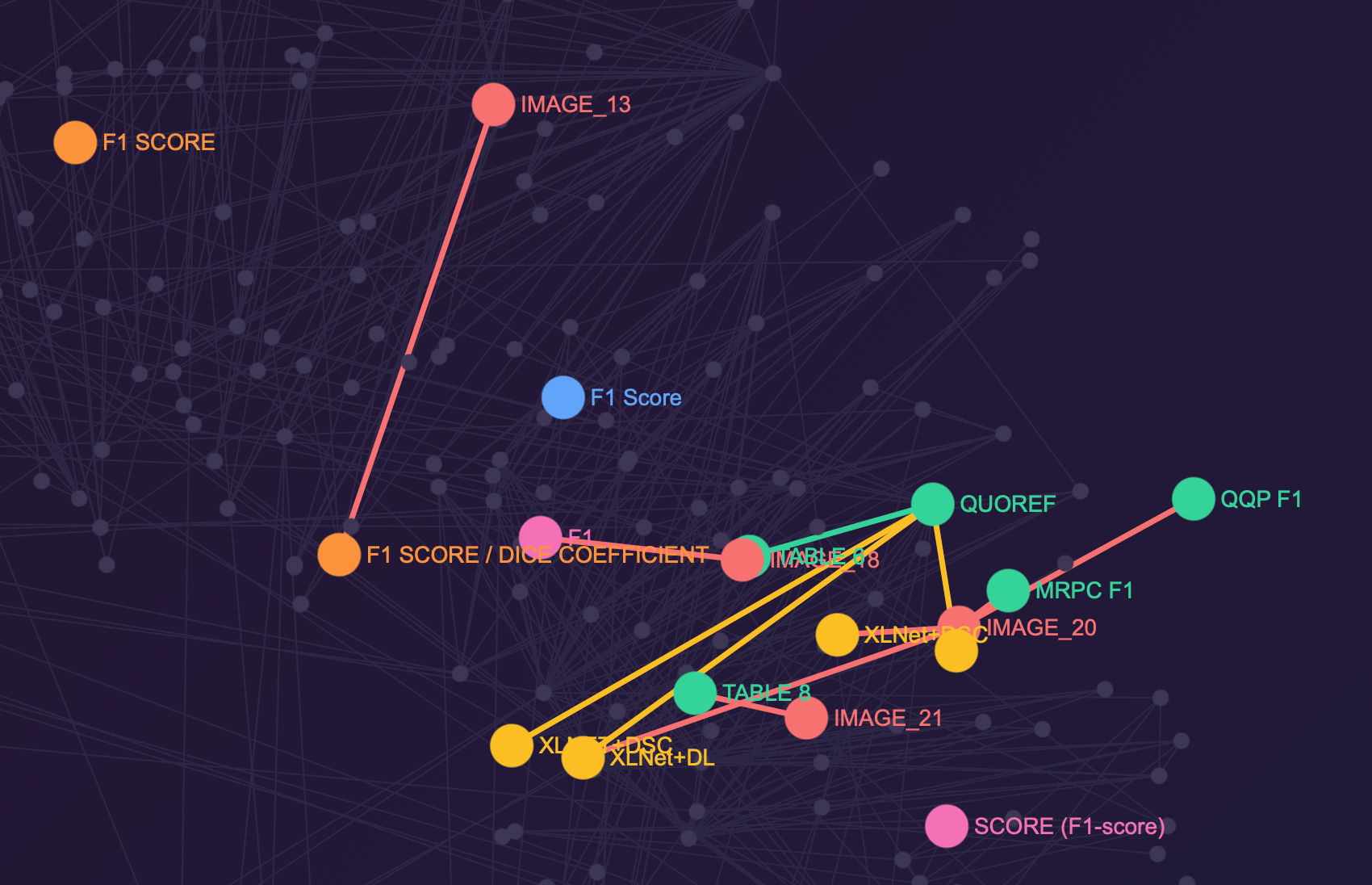}
    \caption{Query-driven extracted local reasoning subgraph (microscopic evidence network formed by multi-hop image-text associations)}
    \Description{Query-specific local MMKG subgraph containing the core entities and relations used for multi-hop reasoning and evidence alignment.}
    \label{fig:sub-mmkg-q3}
\end{figure}

For the given question, the retrieval module performs semantic-similarity-based entity anchoring and topological expansion on the global MMKG, precisely recalling the local reasoning subgraph strongly correlated with ``QuoRef dataset,'' ``XLNet model variants,'' and ``F1 score improvement'' (see Figure~\ref{fig:sub-mmkg-q3}). Core nodes are precisely limited to \nolinkurl{QUOREF}, \nolinkurl{XLNet+DSC}, \nolinkurl{XLNet+DL}, \nolinkurl{TABLE_8}, \nolinkurl{TABLE_6}, and \nolinkurl{F1_SCORE}. This process effectively eliminates irrelevant noise, reducing the large retrieval space to a high-quality multi-hop evidence network connecting task, model, metric, and source.

\subsection{Cross-Modal Evidence Alignment and Logical Reasoning.}

Based on the retrieved local subgraph and retrieval logs, the large language model executes rigorous logical reasoning and conflict resolution along the graph's topological edges:
\begin{enumerate}
    \item \textbf{Task and Baseline Anchoring:} The model first anchors the semantic center at the \nolinkurl{QUOREF} node and baseline model \nolinkurl{XLNet}, then backtracks to its metric source in the original document (anchored to Table 6 text).
    \item \textbf{Variant Topology Expansion:} Following the graph edges from \nolinkurl{XLNet}, the system naturally expands to candidate variant nodes such as \nolinkurl{XLNet+FL} and \nolinkurl{XLNet+DSC}, extracting associated \nolinkurl{F1_SCORE} attributes.
    \item \textbf{Cross-Modal Disambiguation and Branch Downweighting:} During reasoning, the multimodal branch attempts to incorporate visual node \nolinkurl{IMAGE_21} as supplementary evidence. However, through graph relation tracing, the model discovers this image is actually an evaluation table for the QQP task (Table 8), semantically inconsistent with the target task (QuoRef). The system then actively downweights this visual branch, successfully avoiding erroneous cross-modal feature injection and hallucination.
    \item \textbf{Comparative Calculation and Optimal Decision:} Finally, the system performs numerical reasoning on the highly purified aligned evidence. This process can be formalized as solving the following criterion:
    \begin{equation}
        \Delta F1(v) = F1(v, \text{QuoRef}) - F1(\text{XLNet}, \text{QuoRef})
    \end{equation}
    The system extracts the baseline and target values from the graph:
    \begin{align}
        F1(\text{XLNet}, \text{QuoRef}) &= 71.49, \\
        F1(\text{XLNet+DSC}, \text{QuoRef}) &= 72.90.
    \end{align}
    The maximum gain is:
    \begin{equation}
        \Delta F1 = 72.90 - 71.49 = 1.41
    \end{equation}
    Based on this strict calculation chain, the system determines and outputs \nolinkurl{XLNet+DSC} as the model variant with the highest F1 improvement.
\end{enumerate}

\subsection{Interpretability Mechanism Summary.}

Through tracing this reasoning path, it is clear that MMGraphRAG's QA output is not a ``black-box guess'' of large models, but rather built on rigorous provenance across three dimensions: \textbf{Structural Traceability} (ability to drill down from global graph to local subgraph), \textbf{Evidence Traceability} (each generated value can be backward-linked to specific entities, relations, and original image-text blocks), and \textbf{Decision Traceability} (explicit recording of irrelevant branch pruning and downweighting). This architectural design, which converts implicit black-box reasoning into explicit white-box graph traversal, is the core reason why the system maintains extremely low hallucination rates and possesses high interpretability value even in highly complex long-document multimodal scenarios.
\section{Additional Experimental Details: Reproducibility Tables and Protocols}

This appendix collects implementation-level settings that support reproducibility but are too detailed for the main text. These details complement the CMEL, DocQA, and model-scale results reported in the main paper.

\subsection{Method Positioning Across RAG Stages}

Table~\ref{tab:method_positioning_rag_stages} positions MMGraphRAG against representative RAG, graph-reasoning, and multimodal grounding settings. The comparison clarifies which methods build an index from raw documents, which require an external or pre-existing KG, and which stage of the RAG pipeline they primarily target. The table uses open information extraction (OpenIE) and personalized PageRank (PPR) to describe HippoRAG's retrieval stage.

\subsection{SpecLink and CMEL Baseline Configuration}

All CMEL methods use the same candidate scope, image-entity extraction procedure, and final LLM alignment prompt. The shared scope is the boundary-clipped three-chunk neighborhood around the image entity, and all embedding-based comparisons use Stella-en-1.5B-v5. The methods therefore differ only in candidate generation. Table~\ref{tab:cmel_baseline_config} summarizes the operating configuration used in the CMEL comparison.

\begin{table*}[t]
    \centering
        \small
        \setlength{\tabcolsep}{3pt}
        \renewcommand{\arraystretch}{1.08}
        \begin{tabular}{p{0.13\textwidth}p{0.17\textwidth}p{0.13\textwidth}p{0.25\textwidth}p{0.12\textwidth}p{0.10\textwidth}}
        \toprule
        \textbf{Method} & \textbf{Builds raw-doc index?} & \textbf{Prebuilt KG?} & \textbf{Innovation stage} & \textbf{Modality} & \textbf{Supervision} \\
        \midrule
        NaiveRAG & Yes, chunk index & No & Retrieval baseline & Text & Unsupervised \\
        GraphRAG & Yes & No & Index + retrieval & Text & Unsupervised \\
        LightRAG & Yes & No & Retrieval + update & Text & Unsupervised \\
        ToG & No & Yes & Retrieval / reasoning & Text & Unsupervised \\
        RoG & No & Yes & Planning / reasoning & Text & Supervised \\
        GoG & No & Yes, incomplete & Retrieval / generation & Text & Unsupervised \\
        HippoRAG & Auxiliary text graph & No & OpenIE + PPR retrieval & Text & Unsupervised \\
        KG-ViP & No & External KB & Grounded generation & Multimodal & Supervised \\
        CRAG-MM & No, provided sources & Provided sources & Benchmark / evaluation & Multimodal & -- \\
        MMGraphRAG & Yes & No & Multimodal index construction & Multimodal & Unsupervised \\
        \bottomrule
    \end{tabular}
    \caption{Positioning of representative methods and settings across RAG stages. Only MMGraphRAG constructs a multimodal index directly from raw image-text documents without a pre-existing KG.}
    \Description{A table comparing whether representative methods build indexes from raw documents, require prebuilt knowledge graphs, their innovation stage, modality, and supervision.}
    \label{tab:method_positioning_rag_stages}

    \vspace{0.5em}
    \small
    \setlength{\tabcolsep}{4pt}
    \renewcommand{\arraystretch}{1.1}
    \begin{tabular}{p{0.15\textwidth}p{0.48\textwidth}p{0.25\textwidth}}
        \toprule
        \textbf{Method} & \textbf{Key parameters} & \textbf{Candidate selection} \\
        \midrule
        Embedding & threshold $=0.35$ selected on a held-out split & cosine-threshold match \\
        LLM & same visual entity, local context, and text-entity pool & prompt-selected candidates \\
        KMeans & $k=\max(2,\lceil\sqrt{n}\rceil)$ & assigned cluster \\
        DBSCAN & $\epsilon=0.5$; $m=\max(1,\lceil n/10\rceil)$ & assigned cluster \\
        PageRank & weighted relation graph with label-bin partitioning & label-bin cluster \\
        Leiden & modularity partition over the weighted relation graph & community label \\
        SpecLink & LLM-weighted cosine adjacency, $L=D-A$, spectral coordinates, DBSCAN & assigned spectral cluster \\
        \bottomrule
    \end{tabular}
        \caption{CMEL candidate-generation configurations. Here $n$ is the number of nearby text entities and $m$ is the DBSCAN minimum-sample parameter.}
    \Description{A table listing candidate-generation parameters and candidate-selection rules for CMEL comparison methods.}
    \label{tab:cmel_baseline_config}
\end{table*}

For clustering-based methods, the image entity is mapped to a cluster by nearest-neighbor classification over entity descriptions. The reported SpecLink setting uses cosine adjacency reweighted by LLM relation weights, the unnormalized Laplacian $L=D-A$, $k=\max(2,\lceil\sqrt{n}\rceil)$ spectral coordinates, and DBSCAN with the same $\epsilon$ and $m$ values shown in Table~\ref{tab:cmel_baseline_config}. This keeps the final LLM judge identical across methods and isolates candidate-generation quality.

\subsection{SpecLink Parameter Summary}

Table~\ref{tab:speclink_parameter_summary} gives a compact, code-oriented view of the SpecLink fusion component. The candidate pool is defined by a three-chunk neighborhood around the image chunk, clipped at document boundaries. Entity descriptions are embedded with all-MiniLM-L6-v2 for SpecLink, and relation weights scale the cosine adjacency symmetrically before spectral clustering. The multimodal LLM extracts candidate image entities, while the final text LLM is used only for in-candidate alignment, not for candidate generation.

\begin{table}[H]
    \centering
    \small
    \setlength{\tabcolsep}{3pt}
    \renewcommand{\arraystretch}{1.12}
    \begin{tabular}{p{0.36\columnwidth}p{0.52\columnwidth}}
        \toprule
        \textbf{Parameter} & \textbf{Value / setting} \\
        \midrule
        Candidate window & $[i-1,i+1]$, boundary-clipped \\
        Entity embedding & all-MiniLM-L6-v2 sentence encoder \\
        Adjacency & cosine similarity, reweighted by relation importance \\
        Laplacian & $L=D-A$ \\
        Spectral dimension & $k=\max(2,\lceil\sqrt{n}\rceil)$ \\
        Complex handling & modulus of spectral vectors \\
        Clustering & DBSCAN, $\epsilon=0.5$, $m=\max(1,\lceil n/10\rceil)$ \\
        Image-entity assignment & cosine nearest-neighbor, $n_{\mathrm{neighbors}}=3$, top-1 label \\
        Image-entity extraction & multimodal LLM before candidate generation \\
        Final alignment & shared text-LLM prompt over candidate set \\
        \bottomrule
    \end{tabular}
    \caption{SpecLink parameter summary. Here $n$ is the number of nearby text entities and $m$ is the DBSCAN minimum-sample parameter.}
    \Description{A table summarizing the concise values of SpecLink parameters.}
    \label{tab:speclink_parameter_summary}
\end{table}

\subsection{CMEL Evaluation Metric}

CMEL evaluates one-to-one cross-modal entity alignment, so task accuracy directly measures whether each image entity is linked to the correct text entity. Images with no matchable text entity are still included in the decision set: a correct no-match decision is counted as correct, and a wrong match is counted as incorrect. This binary task definition keeps the evaluation focused on the cross-modal linking step rather than on upstream entity extraction.

Under the CMEL assumption, all image entities and candidate text entities have already been produced by Text2Graph and Image2Graph. The isolated linking task therefore has recall fixed by construction and does not use the F1 score, because reporting the F1 score would mix the entity-detection stage with the alignment stage. Accuracy is consequently the cleanest metric for comparing candidate-generation methods under the same candidate scope and final alignment judge.

\subsection{LLM-as-a-Judge Verification Protocol}

The automatic judge is Llama3.1-70B-Instruct. For DocBench, we sample 50 documents with random seed 42, resulting in 251 binary correctness decisions. For MMLongBench, we sample 25 documents with random seed 115, resulting in 222 binary correctness decisions. Human labels are collected as correct/incorrect decisions against the gold answer and then compared with the automatic judge. The agreement reaches 96.0\% on DocBench with Cohen's $\kappa=0.882$, and 94.6\% on MMLongBench with Cohen's $\kappa=0.876$, both within the commonly used ``almost perfect'' agreement range.

The verification artifacts are included in the released resources under \nolinkurl{llm_as_judge_verification}, including extraction statistics and sampled judgement files. The human-verification task is intentionally simple: for each sampled question, the annotator compares the model answer with the short gold answer and assigns a binary correctness label. For example, when the gold answer states that OntoNotes4.0 is a Chinese dataset for named entity recognition, a model answer that identifies OntoNotes4.0 as Chinese is marked correct even if it paraphrases the supporting evidence. Ambiguous cases are re-checked under the same binary rule. Because the sampled items are short-answer factual questions, this protocol reduces the human judgement to direct answer equivalence rather than open-ended preference scoring.

\subsection{Open-Model Selection Rationale}

The experimental pipeline uses three model types: an embedding model, a text LLM, and a multimodal LLM. The goal of the experiments is to evaluate the multimodal graph-construction paradigm rather than to chase the strongest possible leaderboard score, so the backbone models are fixed to representative open models throughout the study. Keeping model versions fixed avoids conflating improvements from the proposed construction method with improvements from changing model backbones during experimentation.

Open models are also important for the target deployment setting. Many document collections in enterprise, legal, medical, and internal-report scenarios cannot be sent to third-party API services, making offline execution a practical requirement. Fixed open-weight models improve reproducibility because other researchers can pin the same versions and rerun the full construction/query pipeline. They also avoid per-token API fees during large-scale indexing and give full control over batching, retries, and system-level optimization. Stronger future backbones may raise absolute scores, but the controlled comparisons in this paper are designed to isolate whether better multimodal graph construction improves retrieval and QA under comparable model settings.

\subsection{Token Accounting}

Tables~\ref{tab:token_accounting_construction} and~\ref{tab:token_accounting_query} report construction-stage and query-stage token usage across the five DocBench document categories. We instrument the pipeline with \texttt{tiktoken} using the Generative Pre-trained Transformer (GPT)-4o encoding, separately count input and output tokens, and report category averages in ktokens, where 1 ktoken is 1,000 tokens. Construction values are per-document averages because the graph index is built once for each document; query values are per-question averages because answering is charged each time a question is processed.

Estimated cost is computed from token counts rather than measured billing records. We use the representative GPT-4o-mini price point of \$0.15 per 1M input tokens and \$0.60 per 1M output tokens. For input count $I_{kt}$ and output count $O_{kt}$ in ktokens, the estimate in US dollars is
\[
\widehat{C}=0.00015 I_{kt}+0.00060 O_{kt}.
\]
The total token column is $I_{kt}+O_{kt}$ and is shown to make the construction/query scale directly comparable.

\begin{table}[t]
    \centering
    \small
    \setlength{\tabcolsep}{3pt}
    \renewcommand{\arraystretch}{1.12}
    \begin{tabular}{lrrrr}
        \toprule
        \textbf{Type} & \textbf{\makecell{Input \\ Tokens}} & \textbf{\makecell{Output \\ Tokens}} & \textbf{\makecell{Total \\ Tokens}} & \textbf{\makecell{Estimated \\ Cost (\$)}} \\
        \midrule
        Aca. & 35.0 & 26.7 & 61.6 & 0.021 \\
        Fin. & 154.9 & 90.0 & 244.8 & 0.077 \\
        Gov. & 87.6 & 59.3 & 147.0 & 0.049 \\
        Law. & 62.4 & 34.8 & 97.2 & 0.030 \\
        New. & 12.5 & 11.4 & 23.9 & 0.009 \\
        \bottomrule
    \end{tabular}
    \caption{Construction-stage token accounting per document. Token counts are in ktokens; costs are estimates computed from input and output token counts.}
    \Description{A table reporting construction-stage input, output, total token counts, and estimated costs by document type.}
    \label{tab:token_accounting_construction}
\end{table}

\begin{table}[t]
    \centering
    \small
    \setlength{\tabcolsep}{3pt}
    \renewcommand{\arraystretch}{1.12}
    \begin{tabular}{lrrrr}
        \toprule
        \textbf{Type} & \textbf{\makecell{Input \\ Tokens}} & \textbf{\makecell{Output \\ Tokens}} & \textbf{\makecell{Total \\ Tokens}} & \textbf{\makecell{Estimated \\ Cost (\$)}} \\
        \midrule
        Aca. & 5.7 & 0.6 & 6.3 & 0.0012 \\
        Fin. & 5.1 & 0.5 & 5.6 & 0.0011 \\
        Gov. & 5.6 & 0.2 & 5.7 & 0.0009 \\
        Law. & 5.1 & 0.3 & 5.4 & 0.0009 \\
        New. & 4.1 & 0.1 & 4.2 & 0.0007 \\
        \bottomrule
    \end{tabular}
    \caption{Query-stage token accounting per question. Token counts are in ktokens; costs are estimates computed from input and output token counts.}
    \Description{A table reporting query-stage input, output, total token counts, and estimated costs by document type.}
    \label{tab:token_accounting_query}
\end{table}

The accounting supports three observations. First, construction dominates a single query but is amortized: documents require 23.9--244.8 ktokens to index, whereas each subsequent question requires only 4.2--6.3 ktokens. Second, construction cost scales mainly with document length and structure rather than the category label itself; finance documents are the most expensive because they are the longest, while single-page news documents are the cheapest. Third, query output is small, averaging only 0.1--0.6 ktokens per question, so query-stage cost is dominated by the retrieved-context input. Once the multimodal graph has been built, the marginal cost of additional questions remains below 0.12 US cents in this estimate.

\end{document}